\documentclass[10pt,twocolumn,letterpaper]{article}

\usepackage[pagenumbers]{cvpr} 

\usepackage{physics}    
\usepackage[normalem]{ulem}  
\usepackage{ragged2e}
\usepackage{wrapfig}
\usepackage{booktabs}
\usepackage{tabularx}
\usepackage{makecell}
\usepackage{siunitx}

\newcolumntype{Y}{>{\RaggedRight\arraybackslash}X} 
\usepackage[dvipsnames]{xcolor}

\definecolor{cvprblue}{rgb}{0.21,0.49,0.74}
\usepackage[pagebackref,breaklinks,colorlinks,citecolor=cvprblue]{hyperref}

\def\paperID{378} 
\def\confName{3DV\xspace}
\def\confYear{2026\xspace}

 \title{Patch-based Representation and Learning for Efficient Deformation Modeling}

\author{
Ruochen Chen
\quad
Thuy Tran
\quad
Shaifali Parashar\vspace{.15cm}\\
\small CNRS, École Centrale de Lyon,
INSA Lyon,
Université Claude Bernard Lyon 1, 
LIRIS, UMR5205, France\\
{\tt\small \{ruochen.chen, dinh-vinh-thuy.tran, shaifali.parashar\}@liris.cnrs.fr}
}

\begin{document}

\twocolumn[{%
  \renewcommand\twocolumn[1][]{#1}  
  \maketitle
  \begin{center}
    \includegraphics[width=\linewidth]{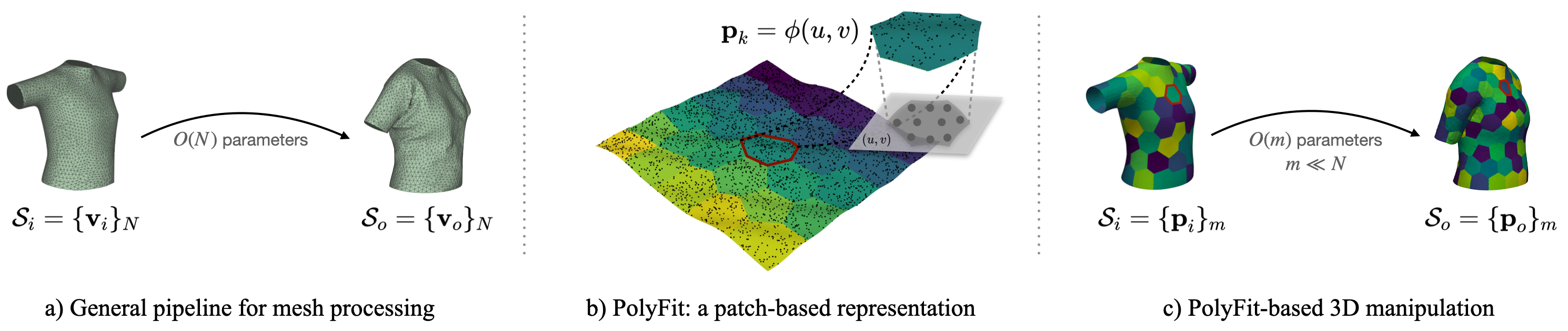}
    \captionof{figure}{\textbf{Patch-based representation and learning.} 
    Against conventional per-vertex parameterizations of mesh deformation (a),
    we present PolyFit (b) which obtains a simplified patch-wise representation by fitting jet functions with limited parameters; consequently simplifying the transformations to modification of patch parameters only (c) and reducing the computational overhead by a large margin. }
    \label{fig:teaser}
  \end{center}
}]%

\begin{abstract}
In this paper, we present a patch-based representation of surfaces, PolyFit, which is obtained by fitting jet functions locally on surface patches. Such a representation can be learned efficiently in a supervised fashion from both analytic functions and real data. Once learned, it can be generalized to various types of surfaces. Using PolyFit, the surfaces can be efficiently deformed by updating a compact set of jet coefficients rather than optimizing per-vertex degrees of freedom for many downstream tasks in computer vision and graphics. We demonstrate the capabilities of our proposed methodologies with two applications: 1) Shape-from-template (SfT): where the goal is to deform the input 3D template of an object as seen in image/video. Using PolyFit, we adopt test-time optimization that delivers competitive accuracy while being markedly faster than offline physics-based solvers, and outperforms recent physics-guided neural simulators in accuracy at modest additional runtime.
2) Garment draping. We train a self-supervised, mesh- and garment-agnostic model that generalizes across resolutions and garment types, delivering up to an order-of-magnitude faster inference than strong baselines.

\end{abstract}
    
\section{Introduction}\label{sec:Introduction}

3D surface deformation is central to many computer vision and graphics applications such as animation~\cite{Grigorev2022}, video editing~\cite{Parashar2019} and medical imaging~\cite{Lamarca2020}, to name but a few. 
 
A common practice is to discretize deformable surfaces as meshes and formulate deformations with per-vertex unknowns, either optimized or predicted. Even when driven by reduced controls or regularizers, the underlying degrees of freedom scale with mesh resolution, which makes optimization and inference costly and hinders cross-resolution generalization. Another possibility is to use parametric representations such as splines~\cite{Bookstein1989} or NURBS~\cite{Piegl1991} to reduce the number of parameters to be estimated; however, such techniques are practical only for simpler geometries. As the geometries grow more complex, the number of parameters required for accurate representation usually explodes; thus defeating the purpose of using a parametric representation as a low-dimensional deformation state. 
Although modern learning-based representations such as AtlasNet~\cite{Groueix2018} and other neural representations~\cite{Yang2023,Park2019,Wang2021,Long2023} alleviate some limitations, their large parameter counts and heavy training typically hinder their usage in surface deformation pipelines that require a compact, controllable state.

In this paper, we present a patch-wise, jet-based representation of surfaces which allows an efficient surface deformation with significantly reduced number of parameters to be estimated. Our proposed representation \emph{PolyFit} divides the surfaces into small patches that are represented by simple jet functions, as seen in~\Cref{fig:teaser}. 
It is learned in a supervised fashion using analytic functions which are cheap to generate. 
If needed, its accuracy can be further improved by fine-tuning with a small number of samples of a given object type. To deform surfaces, 
we directly update the patch-wise $n$-jet coefficients, thereby limiting the number of parameters to be estimated which leads to efficient processing of the deformations. We showcase the efficiency of our proposed methodology with two well-known problems in computer vision and graphics. First, we propose \emph{PolySfT}: a learning-free, polynomial fitting approach to solve SfT~\cite{Salzmann2008,Bartoli2015,Kairanda2022,Stotko2024,Fuentes2022} where the goal is to deform a given 3D template as seen in the images. We show that it performs much faster with competitive accuracy
than the existing best-performing approaches. Second, we propose \emph{OneFit}: a self-supervised polynomial fitting methodology to learn the draping of the garment. Existing methodologies~\cite{Chen2024,Santesteban2022,Bertiche2022} produce mesh-specific or garment-specific solutions. A few exceptions are ~\cite{Grigorev2022, Tiwari2023a, Tiwari2023b}, which can handle multiple garments at various mesh resolutions. However, \cite{Tiwari2023a, Tiwari2023b} are not designed to generate temporally consistent garment deformations as their training process does not incorporate temporal data. In contrast, OneFit is temporally coherent, mesh- and garment-agnostic. Trained on a single garment, OneFit is able to handle 
different inter-class and intra-class garment variations. Moreover, due to its compact representation, it trains faster than existing methods and provides up to an order-of-magnitude faster inference than strong baselines.

\section{Related Work}\label{sec:Related-Work}

\noindent\textbf{Surface representation.}
Parametric chart methods such as AtlasNet~\cite{Groueix2018} learn continuous maps from a 2D domain to 3D surfaces, often trained in a supervised manner on large shape corpora; follow-ups~\cite{Bednarik20,Deng2020} improve efficiency by focusing on local charts. Implicit fields (e.g., SDF/UDF, implicit neural surfaces)~\cite{Park2019,Long2023,Wang2021} model geometry with high capacity but typically require substantial data and compute, and generalization outside the training distribution can be limited. Jets have also been used for local surface fitting~\cite{CAZALS2005121} and differential estimation on point sets~\cite{Ben2020} (e.g., normal/curvature via polynomial jets) and, more recently, neural Jacobian-field~\cite{aigerman2022neural} approaches predict intrinsic Jacobians by supervisedly learning mesh-to-mesh mappings (suitable only for registration or style transfer tasks). Our use of jets differs in both goal and machinery: we make patch-wise $n$-jet coefficients the state variables of deformation with closed-form derivatives, and drive either a learning-free, test-time inverse optimization (PolySfT) or a self-supervised draping model (OneFit). 
Unlike~\cite{Ben2020}, which uses a PointNet~\cite{Qi2017}-based 
weighting module to 
modulate the importance of neighbors to infer point-wise differential quantities, PolyFit performs patch-wise fitting, focusing on efficient deformation modeling rather than per-point geometric property estimation.
Unlike~\cite{aigerman2022neural,Groueix2018}, PolyFit is trained with lightweight synthetic analytic patches (and a small number of garment patches), is local by construction, and generalizes to arbitrary digital 3D surface deformations.

\begin{figure*}[t]
  \centering
  \includegraphics[width=0.93\textwidth]{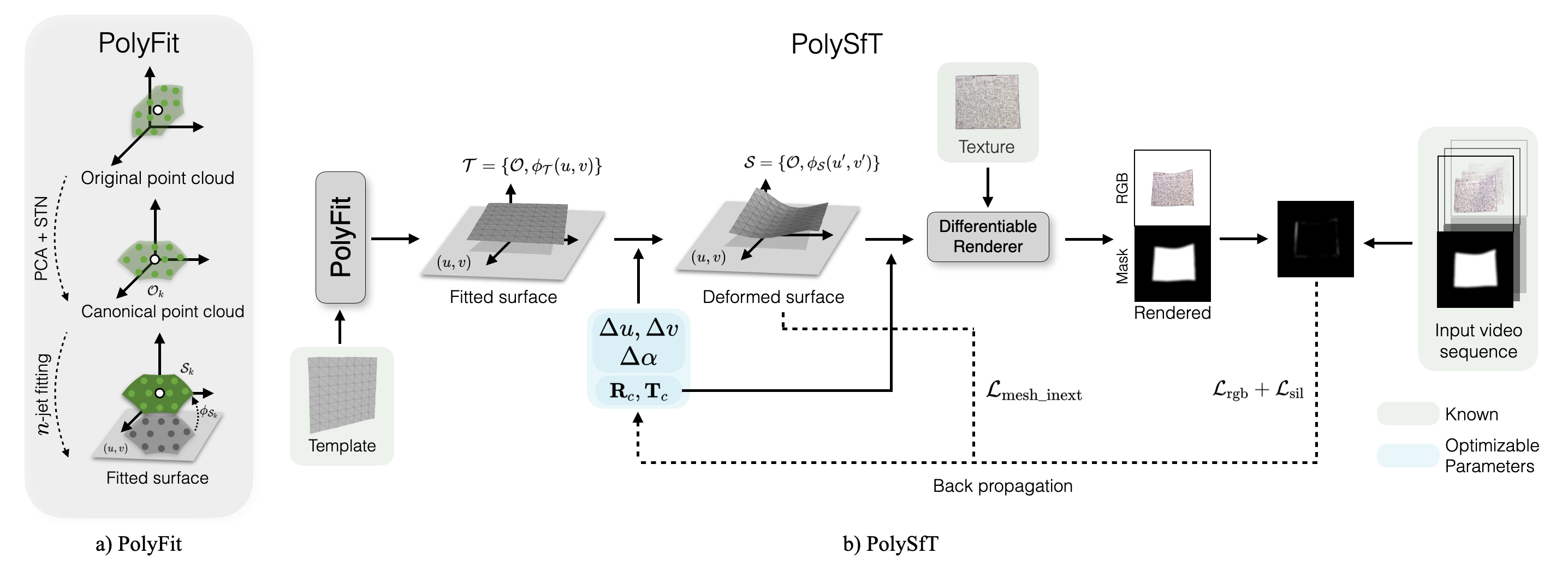}
  \caption{a) \textbf{PolyFit}. It orients the input patch to improve bijectivity with the uv-plane and obtains an analytic representation using n-jet fitting. b) \textbf{PolySfT}. Using PolyFit, the input 3D template is deformed to match input images by estimating the offsets of fitted jet coefficients $\Delta \alpha$, uv displacements $\Delta uv$, as well as a rigid transformation $(\mathbf{R}_c,\mathbf{T}_c)$.}
  \label{fig:polyfit-polysft}
\end{figure*}

\noindent \textbf{Shape-from-template.} Given a known registration between the texture of the template and the images,~\cite{Bartoli2015,Salzmann2008,Chhatkuli2016,Brunet2014,Perriollat2011} compute a unique 3D shape observed in the image, assuming that the object deforms isometrically in a geodesic-preserving fashion, like a piece of paper. Based on this foundational work,~\cite{Ngo2015,Ngo2016,Collins2015,Shetab2024} developed efficient real-time applications. Further advances have been made to incorporate other deformation models such as conformality~\cite{Bartoli2015,Parashar2020}, equiareality~\cite{Parashar2020,Casillas2019}, elasticity~\cite{Malti2013,Malti2015,Malti2017,Haouchine2017,Agudo2015}, ARAP~\cite{Parashar2015} and diffeomorphism~\cite{Ozgur2017}. A major shortcoming of these methods is their inability to handle severe occlusions and capture sharp movements due to the unavailability of high-confidence point correspondences in these scenarios.~\cite{Pumarola2018,Golyanik2018,Shimada2019,Fuentes2022,Fuentes2021} extend SfT to use supervised learning from ground truth data. Recently,~\cite{Sanchez2025} proposed a weakly supervised SfT posing manageable constraints on the input sequence to contain some of the already seen shapes during training. However, all these methods severely degrade under challenging scenarios mentioned above. 

Alternatively,~\cite{Kairanda2022} used physics-based simulation of thin-shell objects~\cite{Liang2019,Narain2012} to deform a 3D template to match input images.~\cite{Stotko2024} used self-supervised learning of physics-based thin-shell simulations~\cite{Santesteban2022,Chen2024} to learn a neural cloth model which operates significantly faster than~\cite{Kairanda2022} although with a degraded performance. 
Our proposed PolySfT uses PolyFit to represent templates with polynomials. It deforms the template by modifying the polynomial parameters to match images 
in a test-time optimization manner, which is significantly more time and memory efficient than offline physics-based simulations where various physical forces are explicitly manipulated.

\noindent \textbf{Garment Draping.} Traditional garment simulation methods rely on computationally expensive but accurate 
physically based cloth simulation
~\cite{Narain2012,Baraff1998,Nealen2006,Macklin2016,Liu2013,Cirio2014}. Advances have been made to reduce the computational complexity of cloth simulation by approximating gradients~\cite{Li2022,Hu2019} for fast computation or adding 3D priors~\cite{Guo2021}  such as point clouds of clothed humans. However, these advances compromise reconstruction quality and make the deployment impractical for virtual try-on systems.
Unlike traditional approaches, modern learning-based methods yield fast inference. Most methods~\cite{Bertiche2020,Patel2020,Santesteban2019,Zhang2021,Wang2019,Lahner2018,Gundogdu2020,Guan2012,Santesteban2021,Pan2022,Vidaurre2020,Shao2023} incorporate a supervised learning approach by using PBS-generated data to learn the relative garment positions with respect to the body. The data generation process is slow and labor intensive which limits the applicability of these methods.
Recently, ~\cite{Bertiche2021,Santesteban2022,Bertiche2022,Chen2024,Grigorev2022} proposed self-supervised learning of garment deformations by converting the physical constraints into optimizable losses to estimate garment positions. Most of these methods learn a mesh-specific model which needs to be retrained for slight changes in the garment topology. To our best knowledge,~\cite{Grigorev2022} is the only exception that uses graph neural networks to learn temporally coherent drapings of several garment meshes. However, the performance decreases while draping meshes with significantly different resolutions from the training. 
In contrast, OneFit transforms garment patches into functions using PolyFit, in order to learn drapings of several garments at all possible mesh resolutions in a self-supervised manner similar to~\cite{Santesteban2022,Chen2024}. Moreover, as compared to mesh-based methods, OneFit is less prone to cloth self-intersections. 

\section{PolyFit}\label{sec:polyfit}

PolyFit allows a jet-based representation on surfaces described with points/meshes. It is possible to represent the entire surface $\mathcal{S}$ with a single function or it can be subdivided using Approximated Centroidal Voronoi Diagrams (ACVD) clustering \cite{Valette2004}, which efficiently constructs uniform tessellations of a given surface area, into desired number of patches. Each patch $k$ is passed into PolyFit which computes orientation $\mathcal{O}_k = (s_k,\mathbf{R}_k,\mathbf{T}_k)$ and a parametric $n$-jet function $\phi_{\mathcal{S}_k}(u,v)$ with respect to a canonical uv space, as seen in~\Cref{fig:polyfit-polysft}(a). Given a set of 3D points $\mathbf{p}_k$ on $\mathcal{S}_k$, PolyFit yields a smooth representation $\mathcal{S}_k:= \{ \mathcal{O}_k,\phi_{\mathcal{S}_k}(u,v) \}$ such that $ \mathbf{p}_k = s_k\mathbf{R}_k\phi_{\mathcal{S}_k}(u,v) + \mathbf{T}_k$. 
Depending on its orientation, projecting $\mathcal{S}_k$ onto a local 2D frame may produce foldovers (overlaps in $(u,v)$), so the surface is no longer a single-valued height graph, breaking the bijectivity of $\phi_{\mathcal{S}_k}$. 
To mitigate this issue, we leverage Principal Component Analysis (PCA) to transform each patch into a canonical space of maximally planar patch representations, which empirically reduces such degeneracies. 
We then use a rigid Spatial Transformer Network (STN)~\cite{Jaderberg2015} parameterized by unit quaternions to refine the orientation and promote a near-bijective height-graph parameterization before $n$-jet fitting (see \Cref{fig:stn_effect} in the supplement for an illustrative example).

Following the explicit representation of surfaces in terms of height function, $z(u,v)$, from a canonical uv space, an $n^{\text{th}}$ order truncated Taylor expansion of $z$ (also known as $n$-jet), is given by $ z(u,v)=\sum_{i=0}^n\sum_{j=0}^i \alpha_{i-j,j}u^{i-j}v^j$.
The combinations of $(\alpha,n)$ allow an analytic representation of various non-trivial geometries, whose $n^{\text{th}}$ order derivatives can be computed precisely. 
Moreover, given sufficient point samples, $z(u,v)$ can be obtained by fitting an $n^{\text{th}}$ order jet in a least squares sense~\cite{Cazals2003}. 
Therefore, canonical representation of surfaces, 
in which every point is parameterized by a diffeomorphism $\phi_{\mathcal{S}_k}:(u,v)\mapsto(u,v,z(u,v))^\top$,  can be oriented using $\mathcal{O}_k=\{s_k, \mathbf{R}_k,\mathbf{T}_k \}$ to fit any smooth surface patch embedded in $\mathbb{R}^3$.

\section{PolySfT}\label{sec:polysft}

PolySfT (see \Cref{fig:polyfit-polysft}(b)) leverages PolyFit to efficiently deform a textured 3D template to match the input images, thereby recovering the 3D shape observed in video. 
We consider a single-patch representation of the template $\mathcal{T}= \{ \mathcal{O},\phi_{\mathcal{T}}(u,v)\}$, where $\mathcal{O} = (s,\mathbf{R},\mathbf{T})$ and $ \phi_\mathcal{T}=(u,v,\sum_{i=0}^n\sum_{j=0}^i \alpha_{i-j,j}u^{i-j}v^j)^\top$. It is deformed using the offset jet coefficients $\Delta \alpha$, uv displacements $(\Delta u, \Delta v)$ as well as a global rotation, $\mathbf{R}_c$ and translation $\mathbf{T}_c$ related to rigid motion of the object. Assuming camera intrinsics are known (a common assumption in SfT), the resulting meshes are converted into RGB and mask images using a differentiable renderer \cite{Laine2020}. These renderings are compared to the target input video sequence by computing pixel-wise RGB and silhouette losses similar to \cite{Kairanda2022}. The gradients of the losses are used to refine the optimizable parameters. 
The reconstructed surface is thus modeled by $\mathbf{R}_c \phi_{\mathcal{S}}(u+\Delta u,v+\Delta v;\alpha+\Delta \alpha)\ + \mathbf{T}_c$ 
followed by transformation given by $\mathcal{O}$, to bring the reconstruction from canonical orientation to real one.

\noindent \textbf{Losses.}
To guide the surface reconstruction, we employ a combination of photometric and geometric losses. Specifically, we adopt an RGB and silhouette loss as defined in \cite{Kairanda2022} to align the reconstructed mesh with the observed imagery. In addition, we apply mesh inextensibility loss to enforce edge-preserving constraints between the deformed and template mesh,
$\mathcal{M}_{\mathcal{P}}$ and $\mathcal{M}_{\mathcal{T}}$ respectively:
\begin{equation}
\label{eq:isometry_mesh}
  \mathcal{L}_{\text{mesh\_inext}} = k_{\text{mi}} \sum_{i=1}^{n_\text{edge}} (e_i(\mathcal{M}_{\mathcal{P}}) - e_i(\mathcal{M}_{\mathcal{T}})) ^ 2
\end{equation}
where $e_i(\cdot)$ denotes $i$-th edge length. 
Furthermore, we introduce a temporal consistency loss defined as follows, which promotes temporal smoothness across frames:
\begin{equation}
\label{eq:tc}
\mathcal{L}_{\text{tc}} = k_{\text{tc}} \frac{1}{W-1} \sum_{t=s}^{s+W-2} \left( \|\delta \alpha_t\|^2 + \|\delta uv_t\|^2 \right).
\end{equation}
where $\delta \alpha_t = \Delta \alpha_{t+1} - \Delta \alpha_t, \quad \delta uv_t = \Delta uv_{t+1} - \Delta uv_t$, and $W$ is the window size.

\section{OneFit}\label{sec:method}

\begin{figure*}[t]
  \centering
  \includegraphics[width=0.9\textwidth]{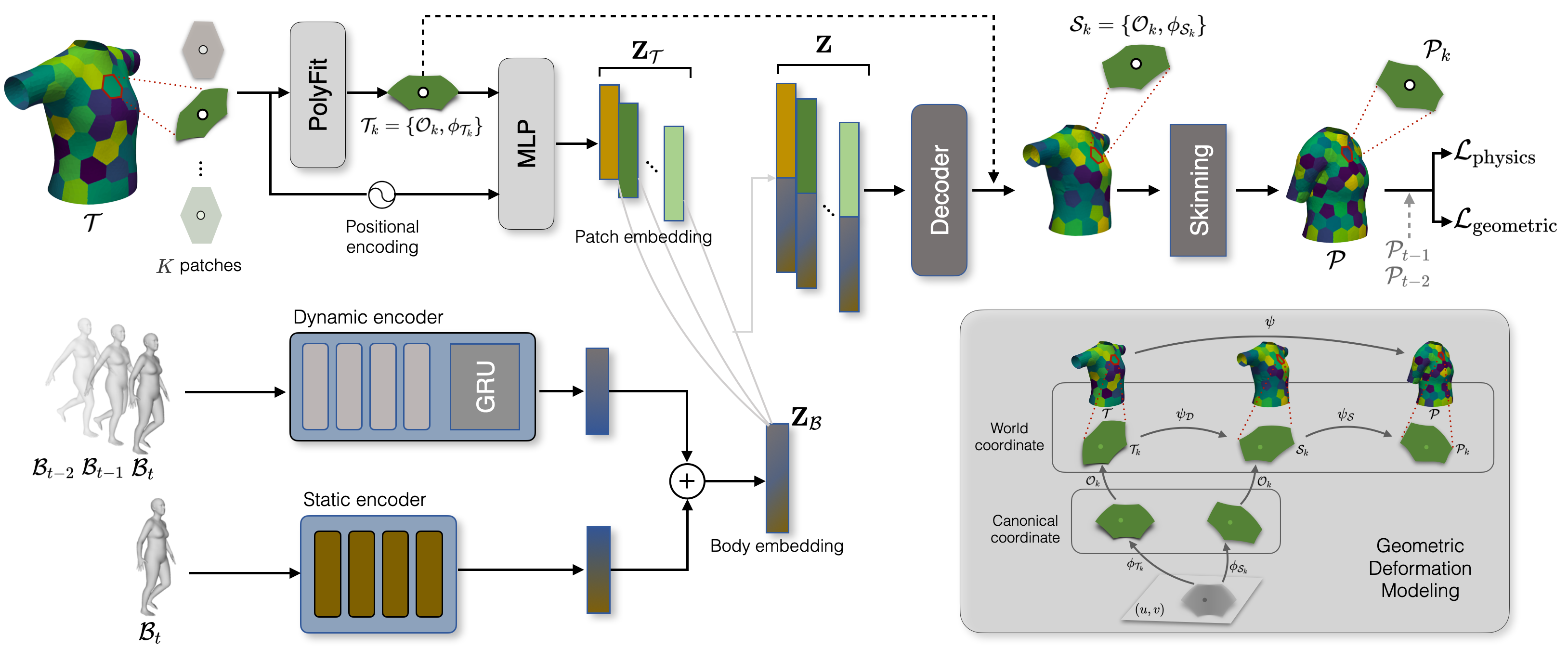}
  \caption{\textbf{OneFit.} 
  It deforms $\mathcal{T}$ isometrically to obtain $\mathcal{P}$ posed on body $\mathcal{B}_t$ by forcing patch boundary consistency, avoiding collisions and maintaining physical equilibria.}
  \label{fig:pipeline-onefit}
\end{figure*}

    
OneFit (see~\Cref{fig:pipeline-onefit}) leverages PolyFit to efficiently simulate garment deformations using self-supervised learning.  The template garment $\mathcal{T}$ is divided into patches using~\cite{Valette2004} which are passed into PolyFit to obtain a smooth patch representation, $\mathcal{T}_k:= \{ \mathcal{O}_k,\phi_{\mathcal{T}_k}(u,v)\}$.


\noindent A \emph{garment patch embedding}, $\mathbf{Z}_{\mathcal{T}_k}$, is generated by passing $\mathcal{T}_k$  along with its positional encoding into the encoder, an MLP with skip connections. The positional encoding, as described in \cite{mildenhall2020}, is applied to each patch to incorporate its center position and its relative offsets from body joints. 

\noindent A \emph{body embedding}, $\mathbf{Z}_{\mathcal{B}}$  is obtained as a concatenation of dynamic and static encoding. To describe joint orientation relative to the parent joint, we follow \cite{Bertiche2022} and adopt 6D descriptors \cite{Zhou2019} concatenated with a unit vector with the unposed direction of gravity. This allows to alleviate the discontinuities in the rotation space presented in axis-angle representation. For the structure of the static and dynamic encoder, we adhere to the framework in~\cite{Bertiche2022}.
The global body pose, $\mathcal{B}(\beta,\theta,\vec{v})$ encapsulates the body shape ($\beta$), the current body pose ($\theta$), and the global velocity of the root joint ($\vec{v}$).

Given $\mathcal{B}(\beta,\theta,\vec{v})$ and $\mathcal{T}$, the network first computes the garment patch and body embeddings, $\mathbf{Z}_{\mathcal{T}_k}$ and $\mathbf{Z}_\mathcal{B}$ respectively. They are then concatenated and fed into a decoder (details in~\Cref{sec:add_details} of the supplement) as $\mathbf{Z}= \text{concatenate}(\mathbf{Z}_{\mathcal{T}_k},\mathbf{Z}_\mathcal{B})$ to predict the patch deformations,  $\mathcal{S}_k:=\{ \mathcal{O}_k,\phi_{\mathcal{S}_k}(u,v) \}$.  The garment deformations are learned by enforcing the physical equilibrium of forces and geometric consistency of template and deformed surface patches posed on the desired body after skinning. This enables a self-supervised, mesh-agnostic, garment-agnostic  learning of the deformations.

\noindent \textbf{Geometric Deformation Modeling.}
As seen in~\Cref{fig:pipeline-onefit}, $\mathcal{T}_k$ is deformed to $\mathcal{S}_k$.
Upon skinning with $\psi_{\mathcal{S}_k}$, we obtain $\mathcal{P}_k$ posed on body $\mathcal{B}_t$. We impose patch deformations to be isometric and enforce the preservation of their first fundamental form in terms of local metric tensors at $\mathcal{T}_k$, $\mathbf{g}_{\mathcal{T}_k}=\mathbf{J}^\top_{\phi_{\mathcal{T}_k}}\mathbf{J}_{\phi_{\mathcal{T}_k}}$  and $\mathcal{P}_k$, $\mathbf{g}_{\mathcal{P}_k} = \mathbf{J}^\top_{\phi_{\mathcal{S}_k}}\mathbf{J}^\top_{\psi_{\mathcal{S}_k}}\mathbf{J}_{\psi_{\mathcal{S}_k}}\mathbf{J}_{\phi_{\mathcal{S}_k}}$.
$\mathbf{J}_{\phi_{\mathcal{T}_k}}$ and $\mathbf{J}_{\phi_{\mathcal{S}_k}}$  can be expressed analytically from the parametric representation obtained in PolyFit. $\mathbf{J}_{\psi_{\mathcal{S}_k}}$ can be analytically calculated from the LBS skinning function~\cite{lin2022fite}.

We impose geometric restrictions on the patch boundaries to maintain consistency. Like~\cite{Chen2024}, we allow local stretchings to avoid collisions. We impose following losses:

\textit{1) Collision.} It penalizes penetration between the body and the garment. For each point, it is given by
\begin{equation}
   \mathcal{L}_{\text{collision}} = k_{\text{c}} \sum_{\text{points}} d_{c}^2,
\end{equation}
 
 where $d_{c}=\max(\epsilon - d(x), 0)$ quantifies the degree of interpenetration. $d(x)$ is the signed distance between the garment vertex and the body surface, and $\epsilon$ is a small positive constant introduced to enhance stability.

\textit{2) Inextensibility.}\label{isometry_loss_function}
To preserve geodesics between the template and draped garment, it enforces metric tensor similarity. It is computed as

\begin{equation}
\label{eq:isometry}
  \mathcal{L}_{\text{inext}} = k_{\text{i}} \frac{1}{KM} \sum_{\mathcal{T}_k\in \mathcal{T}} \sum_{\mathbf{x}\in \mathcal{T}_k} \abs{k_\text{ext}\mathbf{g}_{\mathcal{T}_k}(\mathbf{x})- \mathbf{g}_{\mathcal{P}_k}(\mathbf{x})}
\end{equation}
$M$ is the number of points in each patch 
and $K$ is total number of patches.
$
k_{\text{ext}} = 1 + \min(d_{c}, 0.01)\min(e, 100),
$where $e$ is the current epoch. We first allow network to stabilize and then enforce inextensibility.

\textit{3) Boundary.}
It enforces the connectivity between adjacent patches and is defined as follows:

\begin{equation}
  \mathcal{L}_{\text{boundary}} =   \frac{1}{M_b}\sum_{(i, j)\in \mathcal{B}} \sum_{\text{points}} k_{\text{b}} \| \mathbf{x}_i-\mathbf{x}_j\|^2 +   k_{\text{bn}}  \left(1 - \cos(\theta_n)\right)^2
\end{equation}
where $\mathbf{x}_i$ and $\mathbf{x}_j$ denote boundary points on the adjacent patch of index $i$ and $j$, $M_b$ denotes the total number of adjacent points between all pairs of patches. $\cos(\theta_n)$ represents the cosine similarity between the normals of the $n$-th pair of adjacent points.
It penalizes deviations from perfect parallelism, thus promoting smoother transitions at the boundaries. 
Overall, the geometric losses are given by
\begin{equation}
\label{eq:loss_geometric}
   \mathcal{L}_{\text{geometric}} = \mathcal{L}_{\text{inext}} + \mathcal{L}_{\text{collision}} + \mathcal{L}_{\text{boundary}} +
   \mathcal{L}_{\text{mesh\_inext}}
\end{equation}

$\mathcal{L}_{\text{mesh\_inext}}$, defined in \cref{eq:isometry_mesh}, and $\mathcal{L}_{\text{inext}}$ impose the geodesic preservation constraints at zeroth and first order respectively. It allows the garment deformations to be isometric while taking local body-garment collisions into account.

\noindent \textbf{Physics-based deformation modeling.} 
The physics-based losses incorporate effect of inertia and gravitational forces. The implementation is similar to~\cite{Chen2024} except losses are defined on points instead of mesh vertices. 

\textit{1) Gravity.}
 It incorporates gravity by minimizing the potential energy of the garment, given by
 \begin{equation}
   \mathcal{L}_{\text{gravity}} = \sum_{\text{vertices}}-m g^{\top} \mathbf{x},
 \end{equation}
 where $m$ is the particle mass and $g$ is the gravitational acceleration.

 \textit{2) Inertia.} It incorporates the inertia loss as proposed in ~\cite{Santesteban2022}. It is given by
 \begin{equation}
  \mathcal{L}_{\text{inertia}} = \sum_{\text{vertices}} \frac{1}{2\Delta t^2} m (\mathbf{x}^{[t]}-\mathbf{x}^{[t-1]}- \Delta t v^{[t-1]})^2,
 \end{equation}
 where $\Delta t$ is the simulation time step, $\mathbf{x}^{[t]}$ and $\mathbf{x}^{[t-1]}$ specify the particle's position at times $t$ and $t-1$, respectively. 

Overall, physics-based losses are

\begin{equation}
\label{eq:loss_physics}
   \mathcal{L}_{\text{physics}} =  \mathcal{L}_{\text{inertia}} + \mathcal{L}_{\text{gravity}}
\end{equation}

Together, the losses are given by
\begin{equation}
  \mathcal{L}= \mathcal{L}_{\text{physics}} + \mathcal{L}_{\text{geometric}}
\end{equation}

\section{Experiments}\label{sec:Experiments}

\subsection{PolyFit} \label{sec:polyfit_exp}
We trained PolyFit on point clouds sampled from regular explicit functions (4-jets, trigonometric, Gaussian and Bessels) and on patches sampled from garment meshes in CLOTH3D~\cite{Bertiche2020} dataset, details in \Cref{subsec:Polyfit-training} of the supplement. The training time is about 2 hours.

Note that DeepFit~\cite{Ben2020} and NJF~\cite{aigerman2022neural} are not directly comparable for our setting.
DeepFit performs point-wise jet fitting to estimate local differentials and does not provide a compact patch-level state that can be driven as control variables for surface deformation.
NJF presumes a training set of source–target maps and a global latent code to learn piecewise-linear mesh mappings, supervision that is unavailable in our scenario.

We therefore compare PolyFit against AtlasNet~\cite{Groueix2018}, a learned multi-chart surface generator, for patch fitting.
AtlasNet encodes an input point cloud with PointNet~\cite{Qi2017} and decodes a latent code through \(K\) chart decoders (MLPs), each mapping a 2D parametric domain to a 3D patch; the union of all \(K\) patches forms the reconstruction.
We train the autoencoder variant of AtlasNet with \(K\!\in\!\{5,25,125\}\) on 5{,}000 CLOTH3D garments covering diverse types, and evaluate on six garment templates from~\cite{Santesteban2022}.
For evaluation, we follow the AtlasNet protocol and compute the symmetric Chamfer distance between the reconstruction (concatenating points from all \(K\) charts) and 10,000 points uniformly sampled on the ground-truth template. For both methods, the point clouds are normalized before computing the metric.
We observe that varying \(K\) yields only minor differences in Chamfer distance, so we report the average across \(K\) in \Cref{tab:atlasnet_quan} and provide the full table in \Cref{tab:atlasnet_quan_full} of the supplement.
As summarized in \Cref{tab:atlasnet_quan}, PolyFit consistently attains lower Chamfer distance, demonstrating accurate fitting with analytic, patch-wise representation.
For ablation studies on jet order, the jet-regressor backbone and the training distribution, see \Cref{subsec:Polyfit-experiments} of the supplement.

\begin{table}[htbp]
  \centering
  \resizebox{1\linewidth}{!}{
  \begin{tabular}{@{}lcccccc@{}}
    \toprule
    & Tshirt & Dress & Tank & Top & Shorts & Pants \\
    \midrule
    \textbf{AtlasNet}         & 0.517 & 1.070 & 0.962 & 0.464 & 1.509 & 0.938 \\
    \textbf{PolyFit} (Ours)   & \textbf{0.229} & \textbf{0.168} & \textbf{0.268} & \textbf{0.092} & \textbf{0.372} & \textbf{0.237} \\
    \bottomrule
  \end{tabular}
  }
  \vspace{3pt}
  \caption{Chamfer Distance (multiplied by $10^3$) for patch fitting on six garment templates.}
  \label{tab:atlasnet_quan}
\end{table}

\subsection{PolySfT}
We use adaptive window optimization ($W{=}3$, patience $100$, period $500$; see~\Cref{subsec:polysft_implementation} in the supplement for details). The loss coefficients $k_{\text{mi}}$ and $k_{\text{tc}}$ are set to $0.1$ and $0.05$, respectively. We use Adam optimizer with learning rates $10^{-3}$ for ($\Delta u, \Delta v$), $10^{-2}$ for $\Delta \alpha$, and $10^{-4}$ for both $\mathbf{R}_c$ and $\mathbf{T}_c$.

\begin{figure}[htbp]
  \centering
\includegraphics[width=1\linewidth]{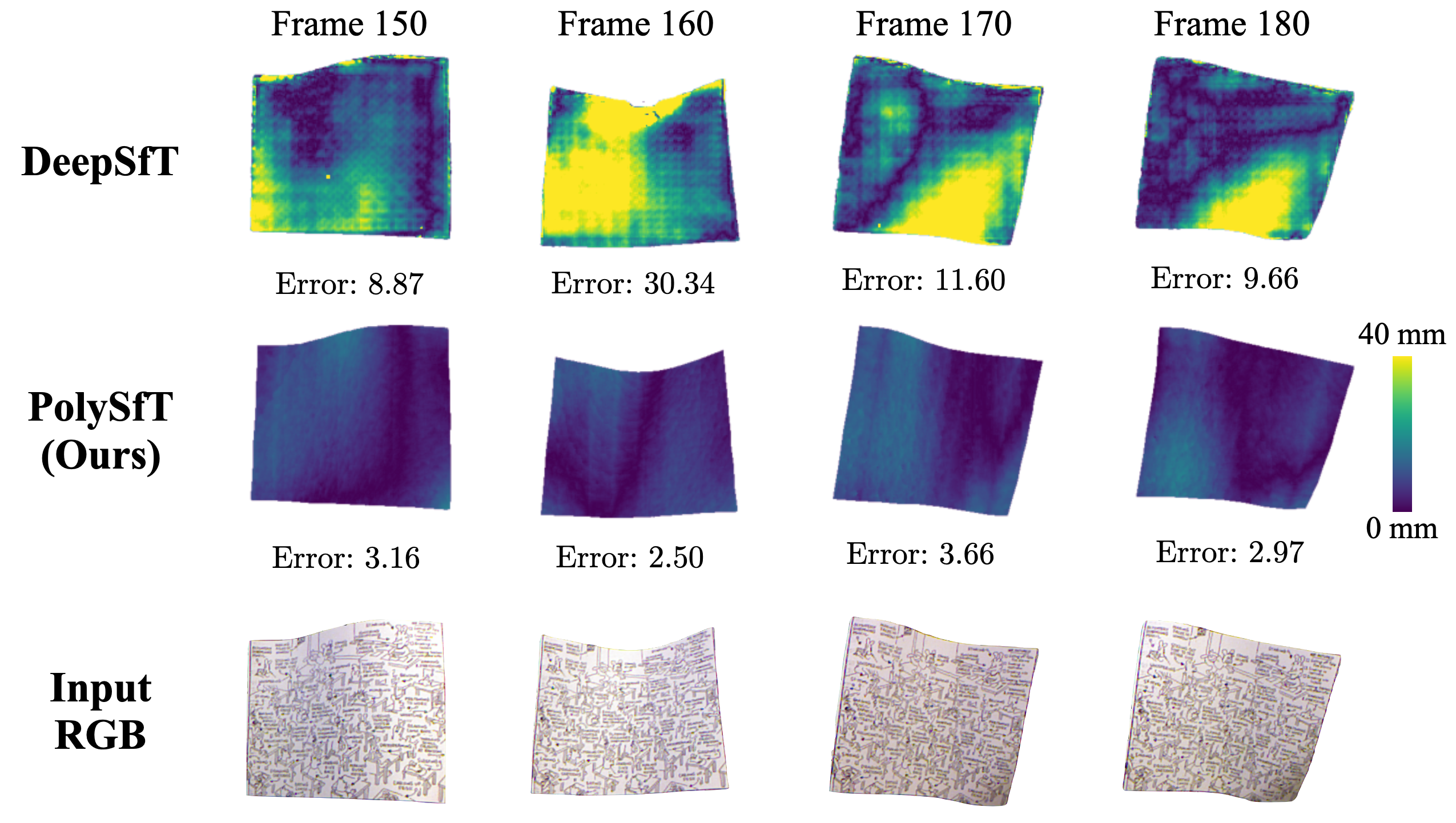}
    \vspace{-5pt}
   \caption{Error map comparison between DeepSfT and Ours on example frames from the Kinect-Paper dataset.}
   \label{fig:sft_qual_sota_kinect_paper}
\end{figure}

We evaluate PolySfT on two real datasets: Kinect-Paper (193 images with ground truth) \cite{Varol2012} and Paper-Bend (71 images without ground truth) ~\cite{Salzmann2007}. ~\Cref{tab:polysft_quantitative} compares quantitative results on Kinect-Paper against traditional SfT (\textsc{SfT})~\cite{Bartoli2015} and supervised SfT methods~\cite{Fuentes2021,Fuentes2022}; PolySfT attains consistently lower errors than these baselines. We did not compute results for $\phi$-SfT \cite{Kairanda2022} 
due to prohibitive runtime and memory requirements on long sequences. 
~\cite{Stotko2024} is designed to work with square meshes only and is therefore not applicable to the Kinect-Paper template. Selected reconstructed frames are shown in \Cref{fig:sft_qual_sota_kinect_paper}. The code for TD-SfT \cite{Fuentes2021} is not publicly available, so we cannot show visual results. 
Additional qualitative results on Paper-Bend are shown in \Cref{fig:sft_qual} of the supplement.

\begin{table}[htbp]
  \centering
  \resizebox{1\linewidth}{!}{
  \begin{tabular}{@{}lcccc@{}}
    \toprule
    & \textbf{\textsc{SfT}} \cite{Bartoli2015} & \textbf{DeepSfT} \cite{Fuentes2022} & \textbf{TD-SfT} \cite{Fuentes2021}  & \textbf{PolySfT} \\
    \midrule
        RMSE (mm)  & 6.17 & 6.97 & 3.37 & \textbf{2.59} \\
    \bottomrule
  \end{tabular}
  }
  \vspace{3pt}
  \caption{\textbf{Kinect-Paper dataset}. Depth RMSE is reported in mm.}
  \label{tab:polysft_quantitative}
\end{table}

In addition, we evaluated PolySfT on synthetic dataset provided by~\cite{Kairanda2022}, which comprises four sequences (S1-S4) of cloth deformations, each containing between 45 and 50 frames. We compute the 3D error $e_{\text{3D}}$  and the average per-vertex angular error $e_{n}$  in degrees, following the definition given in the supplement of \cite{Kairanda2022}. ~\Cref{tab:polysft_quantitative2} shows that our method outperforms PGSfT and \textsc{SfT} on all sequences. It outperforms $\phi$-SfT on S1 and S4, and achieves comparative results on S2 and S3 sequences. Visual comparisons with state-of-the-art methods are shown in~\Cref{fig:qual_phi_sft_syn}. 

We report wall-clock per-frame optimization time averaged over entire sequences. On an NVIDIA V100 GPU, PolySfT runs at $\sim$10s/frame. This is $\sim$270$\times$ faster than $\phi$-SfT ($\sim$2{,}800s/frame) and about $2\times$ slower than PGSfT ($\sim$5s/frame). Despite this gap to PGSfT, PolySfT achieves accurate reconstructions while remaining substantially faster and far more memory-efficient than $\phi$-SfT.

\begin{table}[t]
    \centering
    \resizebox{\linewidth}{!}{%
    \begin{tabular}{@{}lcccccccc@{}}
    \toprule
    & \multicolumn{2}{c}{\textbf{S1}} & \multicolumn{2}{c}{\textbf{S2}} & \multicolumn{2}{c}{\textbf{S3}} & \multicolumn{2}{c}{\textbf{S4}} \\
    & $e_{\text{3D}}$ & $e_{n}$ & $e_{\text{3D}}$ & $e_{n}$ & $e_{\text{3D}}$ & $e_{n}$ & $e_{\text{3D}}$ & $e_{n}$ \\
    \midrule
    \textbf{PGSfT} \cite{Stotko2024} & 0.0298 & 7.780 & 0.0448 & 8.770 & 0.0823 & 21.058 & 0.0919 & 6.885 \\
    \textbf{$\phi$-SfT} \cite{Kairanda2022} & 0.0420 & 11.860 & \textbf{0.0230} & 10.620 & 0.0330 & \textbf{9.120} & 0.0050 & 2.610 \\
    \textbf{SfT} \cite{Bartoli2015} & 0.0328 & 7.275 & 0.0483 & 7.683 & 0.0481 & 14.607 & 0.0232 & 5.165 \\
    \textbf{PolySfT} & \textbf{0.0234} & \textbf{6.337} & \uline{0.0298} & \textbf{4.815} & \textbf{0.0266} & \uline{10.222} & \textbf{0.0026} & \textbf{0.475} \\
    \bottomrule
    \end{tabular}%
    }
    \caption{Results on the $\phi$-SfT synthetic dataset, comparing $e_{\text{3D}}$ and $e_n$ errors across methods for sequences S1 to S4.\textsuperscript{\dag}}
    \label{tab:polysft_quantitative2}
\end{table}

We assess PolySfT's stability by running the optimization beyond 300 iterations, which is our usual checkpoint. We observed the results to stabilize, thus reliably tracking the intended motion, with no visual changes beyond the typical iteration threshold. More details in~\Cref{subsec:polysft_experiments} of the supplement.

\renewcommand{\thefootnote}{\fnsymbol{footnote}}
\footnotetext[2]{Best and second-best results are in bold and with underline, respectively.}
\renewcommand{\thefootnote}{\arabic{footnote}} 

\subsection{OneFit}

We trained OneFit on a set of 6 standard garment templates (tshirt, dress, pants, shorts, long-sleeve top and tank) used in \cite{Santesteban2022}. We utilize the human motion sequences from the AMASS dataset (60 seq., 10K poses) \cite{Mahmood2019}. We then validate the resulting models on unseen garment meshes from CLOTH3D
\cite{Bertiche2020}, where the garment preprocessing steps are described in \Cref{subsec:Garment-preprocessing} of the supplement.

We set the adaptive batch size according to the number of patches of the garment. The learning rate begins at $10^{-3}$ for the first 10 epochs and then reduces to $10^{-4}$ . We set $k_b=5e3$, $k_{mi}=2$, $k_g=1$, $k_c=1$, and $k_i=0.5$. These parameters are fixed for all garments across all experiments. 

We compare OneFit with state-of-the-art self-supervised methods: GAPS~\cite{Chen2024}, SNUG~\cite{Santesteban2022}, NCS~\cite{Bertiche2022} and HOOD~\cite{Grigorev2022}. Except for HOOD, all these methods train mesh-specific models of a single garment. HOOD trains a mesh-based model, but can train a unified model for multiple garments. OneFit trains a mesh-independent model and can learn either a single or a multiple garment network. Furthermore, we can finetune an existing model to a specific garment; thus avoiding from-scratch training. Being mesh-independent, it can generalize to various mesh resolutions.~\Cref{fig:resolution} in the supplement
shows the scalability of OneFit towards various mesh resolutions with a similar inference time. SNUG requires a post-processing to remove garment-body collision artifacts. GAPS learns a body-specific model; thus no post-processing is required. OneFit does not require collision post-processing while dealing with garments and bodies in the training dataset or while dealing with garments which cover the garment-body interactions similar to the training data. However, while dealing with unseen garments, for example trying to drape a full-sleeve tshirt from a model trained on half-sleeve tshirt, some collision artifacts are observed, which can be removed with collision post-processing. 

\begin{figure}[t]
  \centering
\includegraphics[width=1\linewidth]{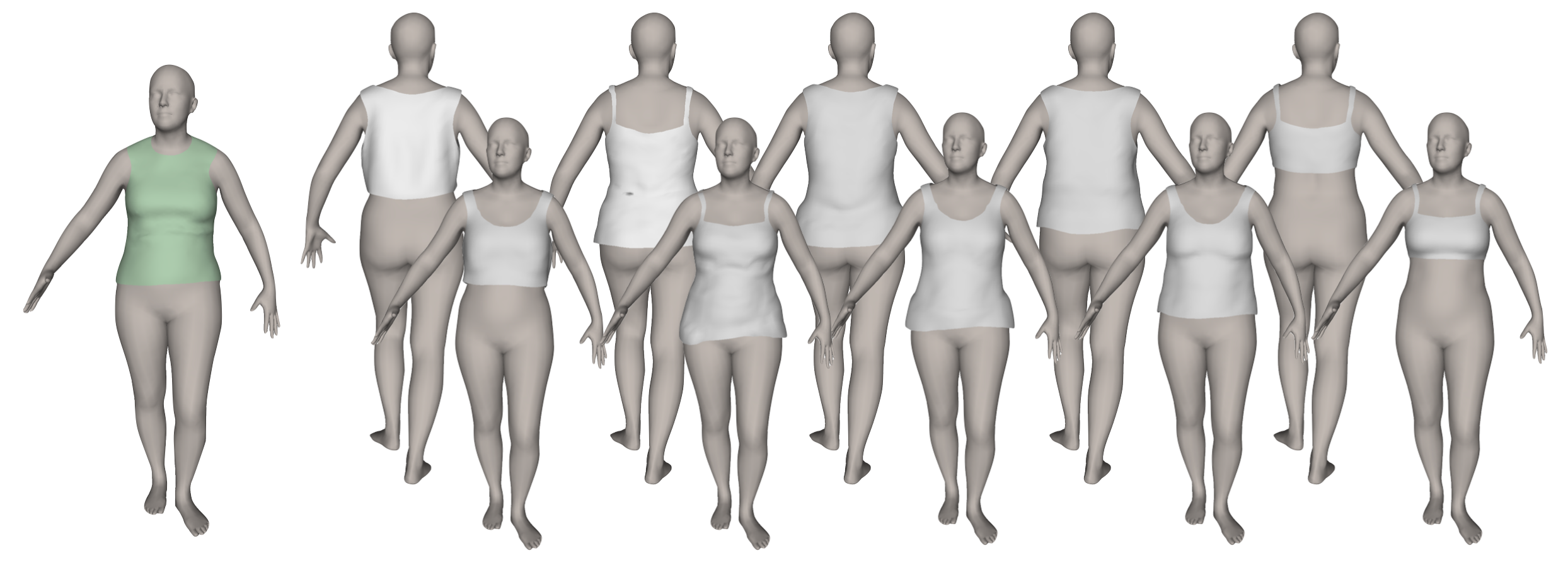}
   \caption{Single garment OneFit under garment intra-class variations. Trained on a Tank top (in green), OneFit is able to drape tank tops of different styles.  }
   \label{fig:onefit-1-generalizability-intra-class}
   \vspace{-5pt} 
\end{figure}

\begin{figure}[htbp]
  \centering
\includegraphics[width=1\linewidth]{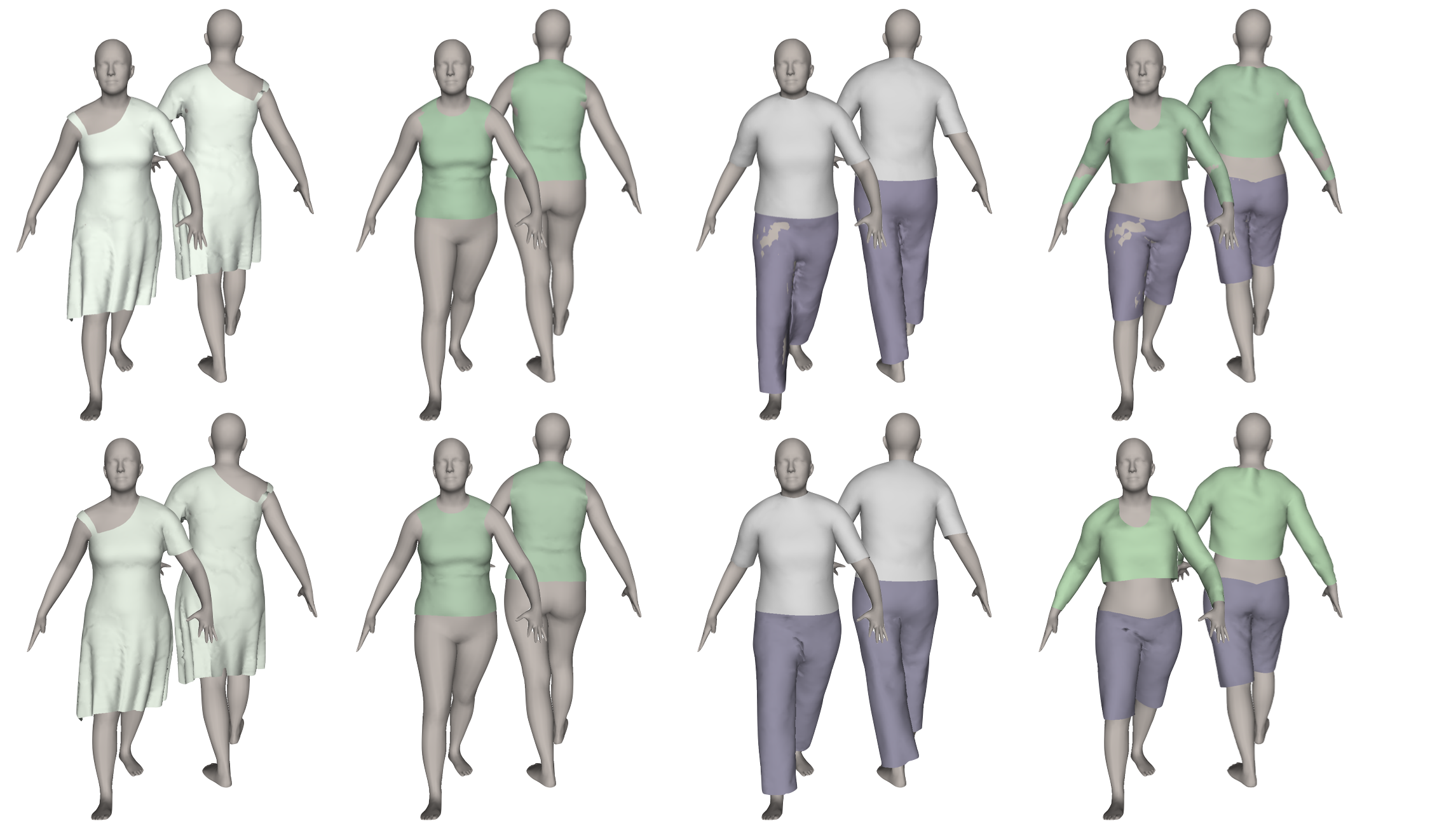}
    \vspace{-5pt}
   \caption{Top: OneFit trained using Dress. Bottom: OneFit trained using a collection of 6 garments.}
   \label{fig:fitting_1-6}
\end{figure}

\noindent \textbf{OneFit as a single garment model.} We test its generalization capabilities.~\Cref{fig:onefit-1-generalizability-intra-class} shows the results of OneFit trained on a Tank top. While it drapes well on the trained garment, it generalizes well to the garments of similar style without a post-processing.  We also test the generalization capabilities of OneFit towards garment inter-class variations.~\Cref{fig:fitting_1-6} (top) shows results of OneFit trained on a dress and tested on various garments. Since it learns garment deformations from small patches, it basically learns localized garment-body interactions which are generally extensible to various garments. Hence, we see a decent drape on tshirt and tank tops. The only artifacts that appear over these garment are due to collisions. Since the network is learned on a dress which does not have arms, it is not trained to be aware of the garment-body interactions in this region which makes the collision artifacts inevitable. A simple post-processing can  remove these artifacts. 
The interesting results in~\Cref{fig:fitting_1-6} (top) are with pants and shorts which are tightly wrapped to the body as opposed to dress. Besides the  collision artifacts, some deformation artifacts are also visible within the area between the legs. Since dress is a loose garment, the network does not witness  tightly-bound garment-body interactions between the legs and produces artifacts. 

\begin{figure}[h]
  \centering
\includegraphics[width=1\linewidth]{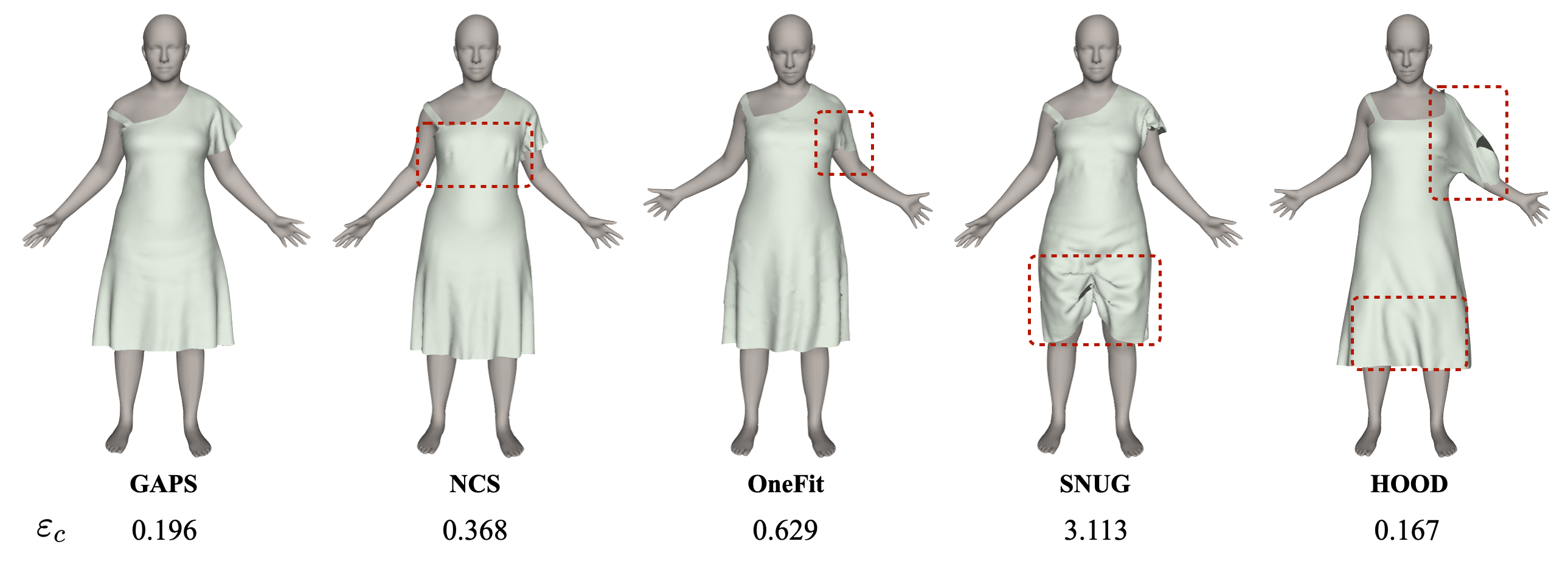}
    \vspace{-10pt}
    \caption{SOTA comparison for OneFit trained on dress. The results on GAPS are reported after post-processing.\textsuperscript{\dag}}
   \label{fig:sota_qual_onefit}
\end{figure}

Loose garments are known to be challenging for most garment draping methods.
~\Cref{fig:sota_qual_onefit} shows that OneFit trained on dress is close to GAPS, the best performing method in this case. All other methods yield noticeable artifacts.

\renewcommand{\thefootnote}{\fnsymbol{footnote}}
\footnotetext[2]{Poses may differ slightly across methods due to differences in the SMPL implementations used by each method.}
\renewcommand{\thefootnote}{\arabic{footnote}}

\begin{figure}[htbp]
  \centering
\includegraphics[width=1\linewidth]{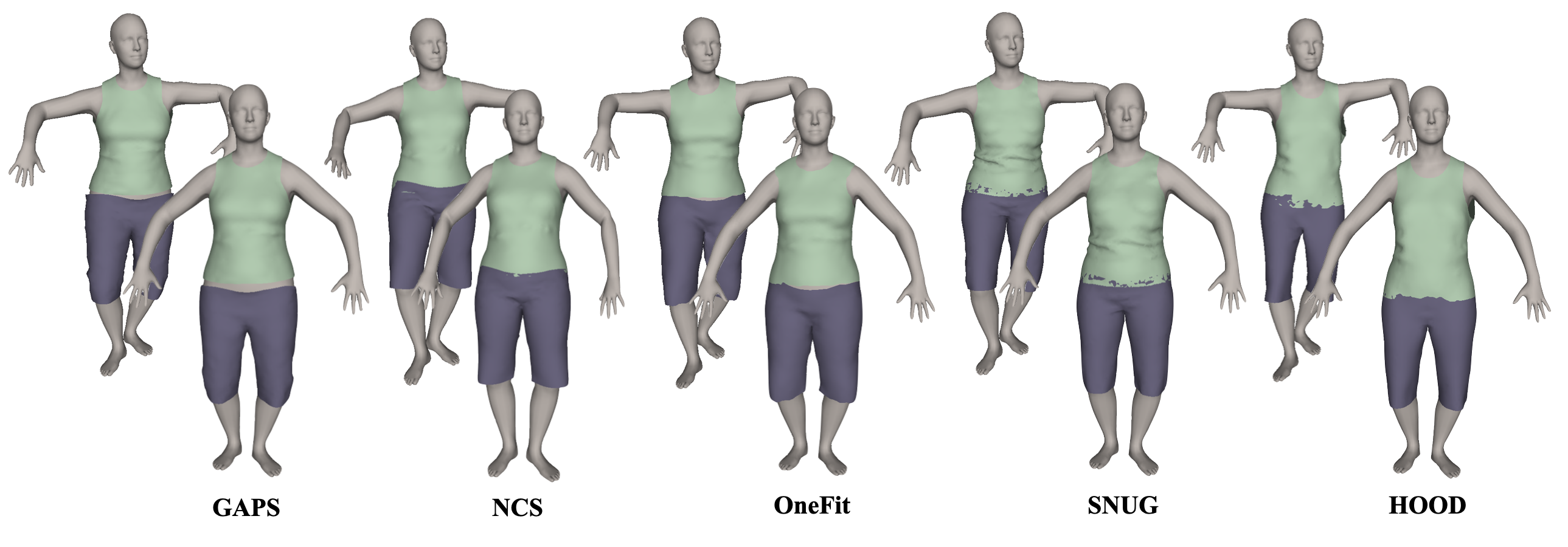}
   \caption{SOTA comparison on tight garments. The results on GAPS are reported after post-processing.\textsuperscript{\dag}}
   \label{fig:sota_qual_b}
\end{figure}

\begin{figure}[htbp]
  \centering
\includegraphics[width=1\linewidth]{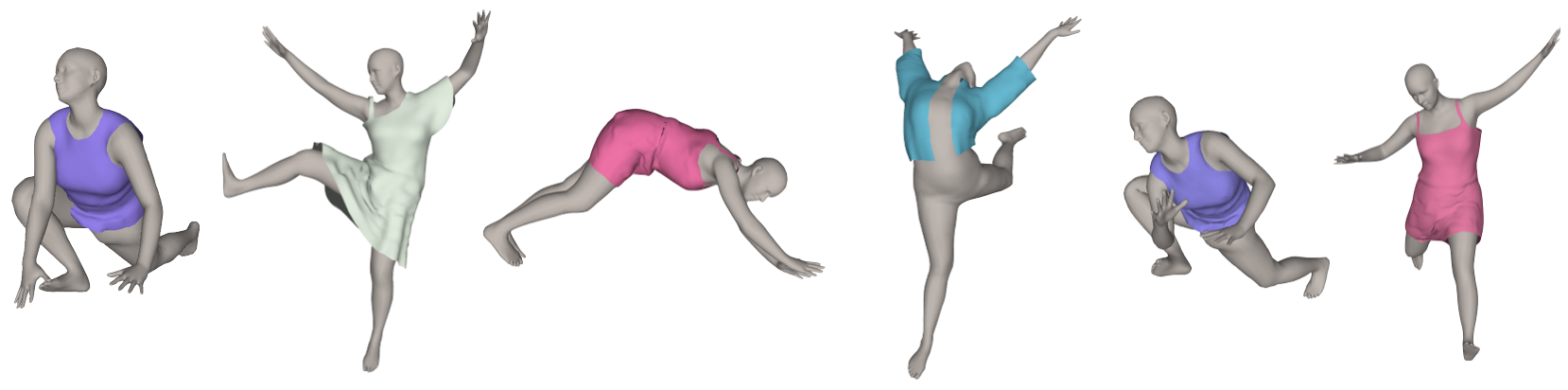}
    \caption{Draping results on challenging poses with diverse garments.}
   \label{fig:extreme_poses}
\end{figure}

\noindent \textbf{OneFit as a multiple garment model.} We trained OneFit jointly on all six garments: tshirt, dress, pants, shorts, long-sleeve top and tank top in order to cover a wide range of body-garment interactions.~\Cref{tab:collision} shows that the $\varepsilon_c$ (\% of vertices under collisions) has drastically reduced as compared to OneFit trained only on dress. 
We also see that training on multiple garments improves the generalizability of OneFit. ~\Cref{fig:fitting_1-6} (bottom) shows better garment drapings on pants and shorts; which demonstrated deformation artifacts in~\Cref{fig:fitting_1-6} (top) under a single garment OneFit.~\Cref{fig:sota_qual_b} shows that OneFit is on par with GAPS, the best performing method in this case. \Cref{fig:extreme_poses} demonstrates the garment-agnostic nature and capability of OneFit to handle extreme poses, including a high-kick dress and a cobra-pose tank-top.

\vspace{-1.5em}

\begin{table}[]
    \centering
    \resizebox{1\linewidth}{!}{%
    \begin{tabular}{@{}lcccccc@{}}
    \toprule
    Model & T-shirt & Dress & Tank & Top & Shorts & Pants \\
    
    \midrule
    \textbf{OneFit} (Dress) & 0.330 & 0.840 & 2.834 & 10.033 & 6.271 & 2.389 \\ 
    \textbf{OneFit} (6 garments) & 0.422 & 0.756 & 0.481 & 1.592 & 1.749 & 1.194 \\     
    \bottomrule
    \end{tabular}
    }
    \caption{$\varepsilon_c$ for various garments. Training on multiple garments improves OneFit's generalizability without requiring any post-processing.}
    \label{tab:collision} 
    
\end{table}
\begin{table}[]
    \centering
    \resizebox{0.85\linewidth}{!}{%
    \begin{tabular}{@{}lcc@{}}
    \toprule
    Model & $\varepsilon_c$ & Training time\\
    \midrule
    \textbf{OneFit} (6 garments) & 2.397 & 8h  \\ 
    \textbf{OneFit} (6 garments) + finetuning  & 1.982 & 1h \\     
    \textbf{OneFit} (jumpsuit) & 1.845 & 3h  \\     
    \bottomrule
    \end{tabular}
    }
    \caption{Fine-tuning vs training OneFit on jumpsuit.}
    \label{tab:fine_tune} 
\vspace{-2pt} 
    
\end{table}


\vspace{-1.5em}

\noindent \textbf{Finetuning OneFit.} Once learned, OneFit can be finetuned to a new garment.~\Cref{tab:fine_tune} compares OneFit trained on multiple garments to drape a new garment, jumpsuit. Almost $~2.5$\% vertices are observed to be under collision which are brought down to less than $2$\% by finetuning this model on jumpsuit for an hour. Training OneFit from scratch achieves a similar performance with $3\times$ more computation. 

\noindent \textbf{Ablation study.} \Cref{tab:abl_study_onefit} shows ablation study on various losses, using a tank top as the test garment.  Besides $\varepsilon_c$, we also compute $\varepsilon_a$ and $\varepsilon_e$ reporting average per-point \% area and edge length change. Losses $\mathcal{L}_\text{mesh\_inext}$ and $\mathcal{L}_\text{inext}$ control the stretchability of garment through zeroth-order (point-based) and first-order (normal-based) metrics; they are both required to minimize size variations while avoiding collisions. $\mathcal{L}_\text{col}$ reduces the amount of body-garment collisions.

\begin{table}[htbp]
  \centering
  \resizebox{0.55\linewidth}{!}{%
    \begin{tabular}{@{}lccc@{}}
      \toprule
      Model & $\varepsilon_e$ & $\varepsilon_a$ & $\varepsilon_c$ \\
      \midrule
      \textbf{OneFit}  & 7.828 & 13.020 & 0.227 \\
      no $\mathcal{L}_\text{mesh\_inext}$   & 13.175 & 24.011 & 0.263 \\
      no $\mathcal{L}_\text{inext}$   & 7.739 & 12.760 & 1.641 \\
      no $\mathcal{L}_\text{col}$   & 8.004 & 13.373 & 0.387 \\
      \bottomrule
    \end{tabular}%
  }
  \caption{Ablation of OneFit training losses.}
  \label{tab:abl_study_onefit}
\end{table}
\vspace{-5pt}

\begin{table}[htbp]
  \centering
  \resizebox{0.7\linewidth}{!}{%
    \begin{tabular}{@{}lcccc@{}}
      \toprule
      & \textbf{SNUG} & \textbf{HOOD} & \textbf{GAPS} & \textbf{OneFit} \\
      \midrule
      Train   & 2 h   & 10 h   & 2-6 h   & 2-8 h   \\
      Runtime & 32.4 ms & 125.5 ms & 5.12 ms & 0.48 ms \\
      \bottomrule
    \end{tabular}%
  }
  \caption{Timing performance.}
  \label{tab:time_perf}
\end{table}

\noindent \textbf{Timing Comparison.}
\Cref{tab:time_perf} reports the timing performance. The mesh specific methods take less time to train but cannot generalize to different topologies. SNUG takes 2 hours to converge for tight garments with less than 10k vertices. GAPS takes 2 hours in the same setting and up to 6 hours for looser garments like dresses. HOOD reports $\sim$10h.
OneFit takes 8 hours for training a multiple garment model on 4 NVIDIA A100 GPUs. 
For runtime, we evaluate on a 2,175-frame CMU sequence. 
HOOD takes the longest runtime. SNUG takes less inference time but is slower than GAPS due to additional collision post-processing. OneFit achieves the fastest runtime; the optional post-processing step (required only while draping garments out of training set) adds only $\sim$3-4ms per frame.

\section{Conclusions}

We introduced PolyFit, a patch-based representation that deforms surfaces via a compact set of jet coefficients. We demonstrated its utility in two applications: PolySfT, a test-time optimization that estimates jet coefficients and local uv shifts so that differentiable renderings match the input images, and OneFit, a self-supervised, mesh- and garment-agnostic neural garment simulation model that generalizes across resolutions and garment types. These results highlight the promise of polynomial, patch-wise representations for efficient deformation modeling/learning.

\noindent \textbf{Limitations.} Each patch in PolyFit is encoded as a single-valued height function; extreme wrinkles, large bulges, or self-occlusions may violate this assumption. We plan to adopt adaptive, curvature/visibility-aware partitioning and enable higher-order jets on demand to preserve fine details. 
Additionally, seam artifacts at patch boundaries may occur under large deformations and are mitigated via a lightweight Laplacian smoothing applied along the boundaries.
In future work, we will develop more effective boundary control that removes this post-processing.

\noindent \textbf{Acknowledgements} This research has received funding from the project RHINO, an ANR-JCJC research grant.

{
    \small
    \bibliographystyle{ieeenat_fullname}
    \bibliography{main}

@String(PAMI = {IEEE Trans. Pattern Anal. Mach. Intell.})

@String(IJCV = {Int. J. Comput. Vis.})

@String(CVPR= {IEEE Conf. Comput. Vis. Pattern Recog.})

@String(ICCV= {Int. Conf. Comput. Vis.})

@String(ECCV= {Eur. Conf. Comput. Vis.})

@String(NIPS= {Adv. Neural Inform. Process. Syst.})

@String(TOG= {ACM Trans. Graph.})

@String(CVPRW= {IEEE Conf. Comput. Vis. Pattern Recog. Worksh.})

@String(CGF  = {Comput. Graph. Forum})

@String(PAMI  = {IEEE TPAMI})

@String(IJCV  = {IJCV})

@String(CVPR  = {CVPR})

@String(ICCV  = {ICCV})

@String(ECCV  = {ECCV})

@String(NIPS  = {NeurIPS})

@String(TOG   = {ACM TOG})

@String(CVPRW= {CVPRW})

@article{Chhatkuli2016,
  title={A stable analytical framework for isometric shape-from-template by surface integration},
  author={Chhatkuli, Ajad and Pizarro, Daniel and Bartoli, Adrien and Collins, Toby},
  journal=PAMI,
  volume={39},
  number={5},
  pages={833--850},
  year={2016},
  publisher={IEEE}
}

@INPROCEEDINGS{Collins2015,
  author={Collins, Toby and Bartoli, Adrien},
  booktitle={2015 IEEE International Symposium on Mixed and Augmented Reality}, 
  title={[POSTER] Realtime Shape-from-Template: System and Applications}, 
  year={2015}
}

@ARTICLE{Ngo2016,
  author={Ngo, Dat Tien and Östlund, Jonas and Fua, Pascal},
  journal=PAMI, 
  title={Template-Based Monocular 3D Shape Recovery Using Laplacian Meshes}, 
  year={2016},
  volume={38},
  number={1},
  pages={172-187}}

@InProceedings{Ngo2015,
author = {Ngo, Dat Tien and Park, Sanghyuk and Jorstad, Anne and Crivellaro, Alberto and Yoo, Chang D. and Fua, Pascal},
title = {Dense Image Registration and Deformable Surface Reconstruction in Presence of Occlusions and Minimal Texture},
booktitle = ICCV,
year = {2015}
}

@article{Shetab2024,
title = {ROBUSfT: Robust real-time shape-from-template, a C++ library},
journal = {Image and Vision Computing},
volume = {141},
pages = {104867},
year = {2024},
author = {Mohammadreza Shetab-Bushehri and Miguel Aranda and Erol Özgür and Youcef Mezouar and Adrien Bartoli},
}

@INPROCEEDINGS{Deng2020,
  author={Deng, Zhantao and Bednařík, Jan and Salzmann, Mathieu and Fua, Pascal},
  booktitle={3DV}, 
  title={Better Patch Stitching for Parametric Surface Reconstruction}, 
  year={2020}
}

@article{Bertiche2022,
    author = {Bertiche, Hugo and Madadi, Meysam and Escalera, Sergio},
    title = {Neural Cloth Simulation},
    year = {2022},
    volume = {41},
    number = {6},
    journal = TOG,
}

@inproceedings{Chen2024,
  title = {GAPS: Geometry-Aware, Physics-Based, Self-Supervised Neural Garment Draping},
  author = {Chen, Ruochen and Parashar, Shaifali and Chen, Liming},
  booktitle = {International Conference on 3D Vision (3DV)},
  year = {2024}
}

@inproceedings{Santesteban2022,
    booktitle = CVPR,
    title = {{SNUG}: {S}elf-{S}upervised {N}eural {D}ynamic {G}arments},
    author = {Santesteban, Igor and Otaduy, Miguel A and Casas, Dan},
    year = {2022}
}

@InProceedings{Salzmann2008,
author="Salzmann, Mathieu
and Moreno-Noguer, Francesc
and Lepetit, Vincent
and Fua, Pascal",
title="Closed-Form Solution to Non-rigid 3D Surface Registration",
booktitle=ECCV,
year="2008"
}

@InProceedings{Stotko2024,
    author    = {Stotko, David and Wandel, Nils and Klein, Reinhard},
    title     = {Physics-guided Shape-from-Template: Monocular Video Perception through Neural Surrogate Models},
    booktitle = CVPR,
    year      = {2024}
}

@inproceedings{Kairanda2022, 
    title={$\phi$-SfT: Shape-from-Template with a Physics-Based Deformation Model}, 
    author={Navami Kairanda and Edith Tretschk and Mohamed Elgharib and Christian Theobalt and Vladislav Golyanik}, 
    booktitle = CVPR, 
    year={2022} 
}

@article{Bartoli2015,
  title={Shape-from-Template},
  author={Bartoli, Adrien and G{\'e}rard, Yan and Chadebecq, Francois and Collins, Toby and Pizarro, Daniel},
  journal=PAMI,
  volume={37},
  number={10},
  pages={2099--2118},
  year={2015},
  publisher={IEEE}
}

@article{Brunet2014,
title = {Monocular template-based 3D surface reconstruction: Convex inextensible and nonconvex isometric methods},
journal = {Computer Vision and Image Understanding},
volume = {125},
pages = {138-154},
year = {2014},
author = {F. Brunet and A. Bartoli and R.I. Hartley}
}

@article{Casillas2019,
title = {Equiareal Shape-from-Template},
journal = {Journal of Mathematical Imaging and Vision},
volume = {61},
number={5},
pages = {607-626},
year = {2014},
author = {Casillas-Perez, David and Pizarro, Daniel and Fuentes-Jimenez, David and Mazo, Manuel and Bartoli, Adrien}
}

@INPROCEEDINGS{Malti2013,
  author={Malti, Abed and Hartley, Richard and Bartoli, Adrien and Kim, Jae-Hak},
  booktitle=CVPR, 
  title={Monocular Template-Based 3D Reconstruction of Extensible Surfaces with Local Linear Elasticity}, 
  year={2013}
}

@INPROCEEDINGS{Malti2015,
  author={Malti, Abed and Bartoli, Adrien and Hartley, Richard},
  booktitle=CVPR, 
  title={A linear least-squares solution to elastic Shape-from-Template}, 
  year={2015}
}

@INPROCEEDINGS{Malti2017,
  author={Malti, Abed and Herzet, Cédric},
  booktitle=CVPR, 
  title={Elastic Shape-from-Template with Spatially Sparse Deforming Forces}, 
  year={2017}
  }

@article{Ozgur2017,
title = {Particle-SfT: A Provably-Convergent, Fast Shape-from-Template Algorithm},
journal = IJCV,
volume = {123},
number={2},
pages = {184-205},
year = {2017},
author = {Özgür, Erol and  Bartoli, Adrien}
}

@INPROCEEDINGS{Haouchine2017,
  author={Haouchine, Nazim and Cotin, Stephane},
  booktitle=CVPR, 
  title={Template-Based Monocular 3D Recovery of Elastic Shapes Using Lagrangian Multipliers}, 
  year={2017}
}

@ARTICLE{Parashar2020,
  author={Parashar, Shaifali and Pizarro, Daniel and Bartoli, Adrien},
  journal=PAMI, 
  title={Local Deformable 3D Reconstruction with Cartan's Connections}, 
  year={2020},
  volume={42},
  number={12},
  pages={3011-3026}
}

@INPROCEEDINGS{Parashar2015,
  author={Parashar, Shaifali and Pizarro, Daniel and Bartoli, Adrien and Collins, Toby},
  booktitle=ICCV, 
  title={As-Rigid-as-Possible Volumetric Shape-from-Template}, 
  year={2015}
}

@INPROCEEDINGS{Agudo2015,
  author={Agudo, Antonio and Moreno-Noguer, Francesc},
  booktitle=CVPR, 
  title={Simultaneous pose and non-rigid shape with particle dynamics}, 
  year={2015}
}

@inproceedings{Pumarola2018,
    title={{Geometry-Aware Network for Non-Rigid Shape Prediction from a Single View}},
    author={A. Pumarola and A. Agudo and L. Porzi and A. Sanfeliu and V. Lepetit and F. Moreno-Noguer},
    booktitle=CVPR,
    year={2018}
}

@InProceedings{Golyanik2018,
author="Golyanik, Vladislav
and Shimada, Soshi
and Varanasi, Kiran
and Stricker, Didier",
title="HDM-Net: Monocular Non-rigid 3D Reconstruction with Learned Deformation Model",
booktitle="Virtual Reality and Augmented Reality",
year="2018",
}

@ARTICLE{Sanchez2025,
  author={Luengo-Sanchez, Sara and Fuentes-Jimenez, David and Losada-Gutierrez, Cristina and Pizarro, Daniel and Bartoli, Adrien},
  journal={IEEE Access}, 
  title={Weakly-Supervised Deep Shape-From-Template}, 
  year={2025},
  volume={13},
  number={},
  pages={22868-22892}}

@INPROCEEDINGS {Shimada2019,
author = { Shimada, Soshi and Golyanik, Vladislav and Theobalt, Christian and Stricker, Didier },
booktitle = CVPRW,
title = {{ IsMo-GAN: Adversarial Learning for Monocular Non-Rigid 3D Reconstruction }},
year = {2019}}

@article{Perriollat2011,
title = {Monocular Template-based Reconstruction of Inextensible Surfaces},
journal = IJCV,
volume = {95},
number={2},
pages = {124-137},
year = {2011},
author = {Perriollat, Mathieu and Hartley, Richard  and  Bartoli, Adrien}
}

@inproceedings{Wang2021,
      title={NeuS: Learning Neural Implicit Surfaces by Volume Rendering for Multi-view Reconstruction}, 
      author={Peng Wang and Lingjie Liu and Yuan Liu and Christian Theobalt and Taku Komura and Wenping Wang},
	  booktitle=NIPS,
      year={2021}
}

@InProceedings{Park2019,
author = {Park, Jeong Joon and Florence, Peter and Straub, Julian and Newcombe, Richard and Lovegrove, Steven},
title = {DeepSDF: Learning Continuous Signed Distance Functions for Shape Representation},
booktitle = CVPR,
year = {2019}
}

@inproceedings{Long2023,
		  title={Neuraludf: Learning unsigned distance fields for multi-view reconstruction of surfaces with arbitrary topologies},
		  author={Long, Xiaoxiao and Lin, Cheng and Liu, Lingjie and Liu, Yuan and Wang, Peng and Theobalt, Christian and Komura, Taku and Wang, Wenping},
		  booktitle=CVPR,
		  year={2023}
		}

@article{Yang2023,
  author       = {Lei Yang and
                  Yongqing Liang and
                  Xin Li and
                  Congyi Zhang and
                  Guying Lin and
                  Alla Sheffer and
                  Scott Schaefer and
                  John Keyser and
                  Wenping Wang},
  title        = {Neural Parametric Surfaces for Shape Modeling},
  journal      = {CoRR},
  volume       = {abs/2309.09911},
  year         = {2023},
  url          = {https://doi.org/10.48550/arXiv.2309.09911},
  doi          = {10.48550/ARXIV.2309.09911},
  eprinttype    = {arXiv},
  eprint       = {2309.09911},
  bibsource    = {dblp computer science bibliography, https://dblp.org}
}

@inproceedings{Groueix2018,
          title={{AtlasNet: A Papier-M\^ach\'e Approach to Learning 3D Surface Generation}},
          author={Groueix, Thibault and Fisher, Matthew and Kim, Vladimir G. and Russell, Bryan and Aubry, Mathieu},
          booktitle=CVPR,
          year={2018}
        }

@article{Lamarca2020,
  title={Defslam: Tracking and mapping of deforming scenes from monocular sequences},
  author={Lamarca, Jose and Parashar, Shaifali and Bartoli, Adrien and Montiel, JMM},
  journal={IEEE Transactions on robotics},
  volume={37},
  number={1},
  pages={291--303},
  year={2020},
  publisher={IEEE}
}

@inproceedings{Parashar2019,
  title={3DVFX: 3D Video Editing using Non-Rigid Structure-from-Motion.},
author={Parashar, Shaifali and Bartoli, Adrien},
booktitle={Eurographics},
year=2019
}

@ARTICLE{Bookstein1989,
  author={Bookstein, F.L.},
  journal=PAMI, 
  title={Principal warps: thin-plate splines and the decomposition of deformations}, 
  year={1989},
  volume={11},
  number={6},
  pages={567-585}
}

@ARTICLE{Piegl1991,
  author={Piegl, L.},
  journal={IEEE Computer Graphics and Applications}, 
  title={On NURBS: a survey}, 
  year={1991},
  volume={11},
  number={1},
  pages={55-71}
}

@inproceedings{Grigorev2022,
author = {Grigorev, Artur and Thomaszewski, Bernhard and Black, Michael J and Hilliges, Otmar}, 
title = {{HOOD}: Hierarchical Graphs for Generalized Modelling of Clothing Dynamics}, 
booktitle = CVPR,
year = {2023},
}

@inproceedings{Bertiche2020,
  title={CLOTH3D: Clothed 3D Humans},
  author={Bertiche, Hugo and Madadi, Meysam and Escalera, Sergio},
  booktitle=ECCV,
  year={2020}
}

@inproceedings{Mahmood2019,
  title = {{AMASS}: Archive of Motion Capture as Surface Shapes},
  author = {Mahmood, Naureen and Ghorbani, Nima and Troje, Nikolaus F. and Pons-Moll, Gerard and Black, Michael J.},
  booktitle = ICCV,
  year = {2019}
}

@article {Santesteban2019,
    journal = CGF,
    title = {{Learning-Based Animation of Clothing for Virtual Try-On}},
    author = {Santesteban, Igor and Otaduy, Miguel A. and Casas, Dan},
    year = {2019},
    volume = {38},
    number ={2},
    pages = {355-366}
}

@article{Wang2019,
author = {Wang, Tuanfeng Y. and Shao, Tianjia and Fu, Kai and Mitra, Niloy J.},
title = {Learning an Intrinsic Garment Space for Interactive Authoring of Garment Animation},
year = {2019},
volume = {38},
number = {6},
journal = TOG,
}

@article{Zhang2021,
author = {Zhang, Meng and Wang, Tuanfeng Y. and Ceylan, Duygu and Mitra, Niloy J.},
title = {Dynamic Neural Garments},
year = {2021},
volume = {40},
number = {6},
journal = TOG,
}

@article {Vidaurre2020,
    journal = CGF,
    title = {{Fully Convolutional Graph Neural Networks for Parametric Virtual Try-On}},
    author = {Vidaurre, Raquel and Santesteban, Igor and Garces, Elena and Casas, Dan},
    year = {2020}
}

@inproceedings{lin2022fite,
  title={Learning Implicit Templates for Point-Based Clothed Human Modeling},
  author={Lin, Siyou and Zhang, Hongwen and Zheng, Zerong and Shao, Ruizhi and Liu, Yebin},
  booktitle={ECCV},
  year={2022}
}

@article{Valette2004,
    author = {Valette, Sébastien and Chassery, Jean-Marc},
    title = {Approximated Centroidal Voronoi Diagrams for Uniform Polygonal Mesh Coarsening},
    journal = CGF,
    volume = {23},
    number = {3},
    pages = {381-389},
    year = {2004}
}

@article{Bertiche2021,
author = {Bertiche, Hugo and Madadi, Meysam and Escalera, Sergio},
title = {PBNS: Physically Based Neural Simulation for Unsupervised Garment Pose Space Deformation},
year = {2021},
volume = {40},
number = {6},
articleno={198},
journal = TOG,
}

@inproceedings {Cazals2003,
booktitle = {Eurographics Symposium on Geometry Processing},
title = {{Estimating Differential Quantities Using Polynomial Fitting of Osculating Jets}},
author = {Cazals, Frédéric and Pouget, Marc},
year = {2003}
}

@inproceedings{Ben2020,
  title={DeepFit: 3D Surface Fitting via Neural Network Weighted Least Squares},
  author={Ben-Shabat, Yizhak and Gould, Stephen},
  booktitle=ECCV,
  year={2020}
}

@inproceedings{Zhou2019,
title={On the Continuity of Rotation Representations in Neural Networks},
author={Zhou, Yi and Barnes, Connelly and Jingwan, Lu and Jimei, Yang and Hao, Li},
booktitle=CVPR,
year={2019}
}

@article {Nealen2006,
journal = CGF,
title = {{Physically Based Deformable Models in Computer Graphics}},
author = {Nealen, Andrew and Mueller, Matthias and Keiser, Richard and Boxerman, Eddy and Carlson, Mark},
year = {2006},
volume = {25},
number = {4}
}

@inproceedings{Baraff1998,
author = {Baraff, David and Witkin, Andrew},
title = {Large Steps in Cloth Simulation},
year = {1998},
booktitle = {Proceedings of the 25th Annual Conference on Computer Graphics and Interactive Techniques},
pages = {43–54}
}

@inproceedings{Macklin2016,
author = {Macklin, Miles and M\"{u}ller, Matthias and Chentanez, Nuttapong},
title = {{XPBD: Position-Based Simulation of Compliant Constrained Dynamics}},
year = {2016},
booktitle = {Proceedings of the 9th International Conference on Motion in Games},
}

@article{Cirio2014,
author = {Cirio, Gabriel and Lopez-Moreno, Jorge and Miraut, David and Otaduy, Miguel A.},
title = {Yarn-Level Simulation of Woven Cloth},
year = {2014},
volume = {33},
number = {6},
journal = TOG,
}

@InProceedings{Qi2017,
  author    = {Charles R. Qi and Hao Su and Kaichun Mo and Leonidas J. Guibas},
  title     = {{PointNet: Deep Learning on Point Sets for 3D Classification and Segmentation}},
  booktitle = CVPR,
  year      = {2017},
}

@article{Gundogdu2020,
  author = {Gundogdu, Erhan and Constantin, Victor and Parashar, Shaifali and Seifoddini, Amrollah and Dang, Minh and Salzmann, Mathieu and Fua, Pascal},
  title = {{Garnet++: Improving Fast and Accurate Static 3D Cloth Draping by Curvature Loss}},
  journal = PAMI,
  volume = 44,
  number = 1,
pages = {181--195},
  year = 2020
}

@inproceedings{Patel2020,
        title = {{TailorNet: Predicting Clothing in 3D as a Function of Human Pose, Shape and Garment Style}},
        author = {Patel, Chaitanya and Liao, Zhouyingcheng and Pons-Moll, Gerard},
        booktitle = CVPR,
        year = {2020},
    }

@inproceedings{Bednarik20,
   title = {Shape Reconstruction by Learning Differentiable Surface Representations},
   author = {Bednarik, Jan and Parashar, Shaifali and Gundogdu, Erhan and Salzmann, Mathieu and Fua, Pascal},
   booktitle = CVPR,
   year = {2020}
}

@article{Liu2013,
author = {Liu, Tiantian and Bargteil, Adam W. and O'Brien, James F. and Kavan, Ladislav},
title = {Fast Simulation of Mass-Spring Systems},
year = {2013},
volume = {32},
number = {6},
journal = TOG,
pages = {1-7}
}

@inproceedings{Lahner2018,
	title = {{DeepWrinkles}: {Accurate} and {Realistic} {Clothing} {Modeling}},
	booktitle = ECCV,
	author = {Lähner, Zorah and Cremers, Daniel and Tung, Tony},
	year = {2018}
}

@article{Narain2012,
	title = {Adaptive Anisotropic Remeshing for Cloth Simulation},
	volume = {31},
	number = {6},
	journal = TOG,
	author = {Narain, Rahul and Samii, Armin and O'Brien, James F.},
	year = {2012},
	pages = {147:1--10}
}

@INPROCEEDINGS{Guo2021,
  author={Guo, Jingfan and Li, Jie and Narain, Rahul and Park, Hyun Soo},
  booktitle=CVPR, 
  title={Inverse Simulation: Reconstructing Dynamic Geometry of Clothed Humans via Optimal Control}, 
  year={2021}
}

@article{Li2022,
author = {Li, Yifei and Du, Tao and Wu, Kui and Xu, Jie and Matusik, Wojciech},
title = {DiffCloth: Differentiable Cloth Simulation with Dry Frictional Contact},
year = {2022},
volume = {42},
number = {1},
journal = TOG,
}

@inproceedings{Liang2019,
author = {Liang, Junbang and Lin, Ming C. and Koltun, Vladlen},
title = {Differentiable cloth simulation for inverse problems},
year = {2019},
booktitle = NIPS,
}

@INPROCEEDINGS{Hu2019,
  author={Hu, Yuanming and Liu, Jiancheng and Spielberg, Andrew and Tenenbaum, Joshua B. and Freeman, William T. and Wu, Jiajun and Rus, Daniela and Matusik, Wojciech},
  booktitle={ICRA}, 
  title={ChainQueen: A Real-Time Differentiable Physical Simulator for Soft Robotics}, 
  year={2019},
  }

@article{Guan2012,
author = {Guan, Peng and Reiss, Loretta and Hirshberg, David A. and Weiss, Alexander and Black, Michael J.},
title = {DRAPE: DRessing Any PErson},
year = {2012},
volume = {31},
number = {4},
articleno={35},
journal = TOG,
}

@inproceedings {Santesteban2021,
    booktitle = CVPR,
    title = {{Self-Supervised Collision Handling via Generative 3D Garment Models for Virtual Try-On}},
    author = {Santesteban, Igor and Thuerey, Nils and Otaduy, Miguel A and Casas, Dan},
    year = {2021}
}

@inproceedings{Pan2022,
   title={Predicting Loose-Fitting Garment Deformations Using Bone-Driven Motion Networks},
   booktitle=SIGGRAPH,
   author={Pan, Xiaoyu and Mai, Jiaming and Jiang, Xinwei and Tang, Dongxue and Li, Jingxiang and Shao, Tianjia and Zhou, Kun and Jin, Xiaogang and Manocha, Dinesh},
   year={2022}
}

@inproceedings{Mildenhall2020,
  title={NeRF: Representing Scenes as Neural Radiance Fields for View Synthesis},
  author={Ben Mildenhall and Pratul P. Srinivasan and Matthew Tancik and Jonathan T. Barron and Ravi Ramamoorthi and Ren Ng},
  year={2020},
  booktitle={ECCV},
}

@ARTICLE{Varol2012,
  author={Varol, Aydin and Shaji, Appu and Salzmann, Mathieu and Fua, Pascal},
  journal=PAMI, 
  title={Monocular 3D Reconstruction of Locally Textured Surfaces}, 
  year={2012},
  volume={34},
  number={6},
  pages={1118-1130}}

@INPROCEEDINGS{Salzmann2007,
  author={Salzmann, Mathieu and Hartley, Richard and Fua, Pascal},
  booktitle=ICCV, 
  title={Convex Optimization for Deformable Surface 3-D Tracking}, 
  year={2007}
}

@article{Fuentes2022,
title = {Deep Shape-from-Template: Single-image quasi-isometric deformable registration and reconstruction},
journal = {Image and Vision Computing},
volume = {127},
pages = {104531},
year = {2022},
issn = {0262-8856},
doi = {https://doi.org/10.1016/j.imavis.2022.104531},
url = {https://www.sciencedirect.com/science/article/pii/S0262885622001603},
author = {David Fuentes-Jimenez and Daniel Pizarro and David Casillas-Pérez and Toby Collins and Adrien Bartoli},
}

@ARTICLE{Fuentes2021,
  author={Fuentes-Jimenez, David and Pizarro, Daniel and Casillas-Perez, David and Collins, Toby and Bartoli, Adrien},
  journal={IEEE Access}, 
  title={Texture-Generic Deep Shape-From-Template}, 
  year={2021},
  volume={9},
  number={},
  pages={75211-75230},
  doi={10.1109/ACCESS.2021.3082011}}

@InProceedings{Tiwari2023a,
    author    = {Tiwari, Lokender and Bhowmick, Brojeshwar},
    title     = {GarSim: Particle Based Neural Garment Simulator},
    booktitle = {Proceedings of the IEEE/CVF Winter Conference on Applications of Computer Vision (WACV)},
    month     = {January},
    year      = {2023},
    pages     = {4472-4481}
}

@InProceedings{Tiwari2023b,
    author    = {Tiwari, Lokender and Bhowmick, Brojeshwar and Sinha, Sanjana},
    title     = {GenSim: Unsupervised Generic Garment Simulator},
    booktitle = {Proceedings of the IEEE/CVF Conference on Computer Vision and Pattern Recognition (CVPR) Workshops},
    month     = {June},
    year      = {2023},
    pages     = {4169-4178}
}

@inproceedings{Shao2023,
  author = {Shao, Yidi and Loy, Chen Change and Dai, Bo},
  title = {Towards Multi-Layered {3D} Garments Animation},
  booktitle = ICCV,
  year = {2023}
}

@aInProceedings{aigerman2022neural,
  title={Neural Jacobian Fields: Learning Intrinsic Mappings of Arbitrary Meshes},
  author={Aigerman, Noam and Gupta, Kunal and Kim, Vladimir G and Chaudhuri, Siddhartha and Saito, Jun and Groueix, Thibault},
  booktitle={SIGGRAPH},
  year={2022}
}

@article{Laine2020,
  title   = {Modular Primitives for High-Performance Differentiable Rendering},
  author  = {Laine, Samuli and Hellsten, Janne and Karras, Tero and Seol, Yeongho and Lehtinen, Jaakko and Aila, Timo},
  journal = {ACM Transactions on Graphics},
  year    = {2020},
  volume  = {39},
  number  = {6}
}

@inproceedings{Jaderberg2015,
author = {Jaderberg, Max and Simonyan, Karen and Zisserman, Andrew and Kavukcuoglu, Koray},
title = {Spatial transformer networks},
year = {2015},
publisher = {MIT Press},
address = {Cambridge, MA, USA},
booktitle = {Proceedings of the 29th International Conference on Neural Information Processing Systems - Volume 2},
pages = {2017–2025},
numpages = {9},
location = {Montreal, Canada},
series = {NIPS'15}
}

@article{CAZALS2005121,
title = {Estimating differential quantities using polynomial fitting of osculating jets},
journal = {Computer Aided Geometric Design},
volume = {22},
number = {2},
pages = {121-146},
year = {2005},
author = {F. Cazals and M. Pouget}
}
}
\clearpage
\setcounter{page}{1}
\maketitlesupplementary
\subsection{PolyFit: Training and Implementation}\label{subsec:Polyfit-training}

\noindent \textbf{Training dataset.} 
To support the training of the rotation correction block in PolyFit, we created a dataset consisting of point cloud patches, generated by combining four families of functions, including jet, trigonometric, Gaussian and Bessel. The four families of functions are:

\textit{1) $4$-jet:} 
$  f(u, v)=\sum_{i=0}^4\sum_{j=0}^i \alpha_{i-j,j}u^{i-j}v^j$ 

\textit{2) Trigonometric:}  
$  T(u, v) = \alpha \cos(\theta \sqrt{u^2 + v^2}) $ 

\textit{3) Gaussian:} 
$  G(u, v) = \alpha \exp\left(-\frac{(u - u_0)^2 + (v - v_0)^2}{2\sigma^2}\right) $ 

\textit{4) Bessel:}
$  B(u, v) = \alpha J_0\left(k \sqrt{(u - u_0)^2 + (v - v_0)^2}\right) $ 

where $ \alpha \in [-0.5, 0.5] $  ,  $ \theta \in [\pi, 2\pi]$ , $ \sigma \in [0.5, 1] $ and $k=5$. Here, $J_0$ denotes the Bessel function of the first kind of order 0. Using $(u, v) \in [-1, 1]$, we sum the outputs from the four functions and train the PolyFit in a supervised way, by minimizing the height discrepancies between the original and the fitted surface points. We further add patches extracted from CLOTH3D~\cite{Bertiche2020} training dataset. The garment meshes are first subdivided four times to achieve a dense mesh. ACVD~\cite{Valette2004} is applied to the refined mesh, clustering the vertices into $k$ patches according to the surface area. Specifically, the number of patches is given by 
$\max\left(100, \min\left(400, \left\lfloor\frac{A}{0.008}\right\rfloor\right)\right)$,
where $A$ denotes the area of the mesh. We extracted $100$k patches and computed ground truth normals from their corresponding meshes.

\begin{figure}[h]
  \centering
\includegraphics[width=1\linewidth]{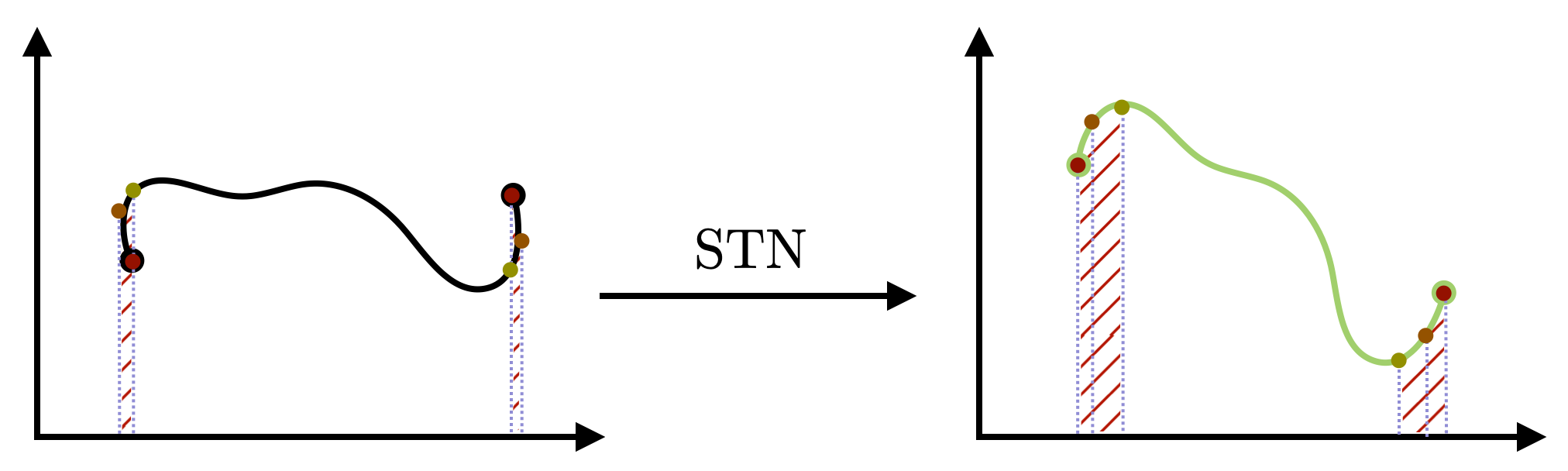}
   \caption{Effect of STN canonicalization. It promotes a near-bijective parameterization (1-D section shown).}
   \label{fig:stn_effect}
\end{figure}

\noindent \textbf{Training details.} The batch size is set to 512 and the learning rate is set to 0.001. For every patch, we perform a preprocessing step including normalization, basis extraction and coordinate frame transformation, as depicted in \cite{Ben2020}. \Cref{fig:stn_effect} illustrates the benefit of using STN correction module, which refines the orientation of the given input point cloud and promotes a near-bijective height-graph parameterization before $n$-jet fitting.

\subsection{PolyFit: Experiments}\label{subsec:Polyfit-experiments}

\paragraph{Comparison with AtlasNet.}
We report per-template Chamfer distances for AtlasNet and PolyFit across \(K\!\in\!\{5,25,125\}\) learned charts.
For training, we use square (patch) as template type, the number of sampled points is set to 10,000. The point clouds are normalized before computing the metric.
As seen in \Cref{tab:atlasnet_quan_full}, varying \(K\) leads to only minor CD changes (the total point budget is fixed), while PolyFit attains consistently lower errors on all templates.

\begin{table}[htbp]
  \centering
  \resizebox{1\linewidth}{!}{
  \begin{tabular}{@{}lcccccc@{}}
    \toprule
    & Tshirt & Dress & Tank & Top & Shorts & Pants \\
    \midrule
    AtlasNet ($K=5$)    & 0.531 & 1.039 & 0.939 & 0.481 & 1.508 & 0.963 \\
    AtlasNet ($K=25$)   & 0.490 & 1.060 & 0.942 & 0.420 & 1.471 & 0.992 \\
    AtlasNet ($K=125$)  & 0.531 & 1.111 & 1.005 & 0.490 & 1.547 & 0.858 \\
    PolyFit (Ours)      & \textbf{0.229} & \textbf{0.168} & \textbf{0.268} & \textbf{0.092} & \textbf{0.372} & \textbf{0.237} \\
    \bottomrule
  \end{tabular}
  }
  \vspace{3pt}
  \caption{Chamfer Distance (multiplied by $10^3$) for patch fitting on six garment templates.}
  \label{tab:atlasnet_quan_full}
\end{table}

\begin{figure}[h]
  \centering
\includegraphics[width=1\linewidth]{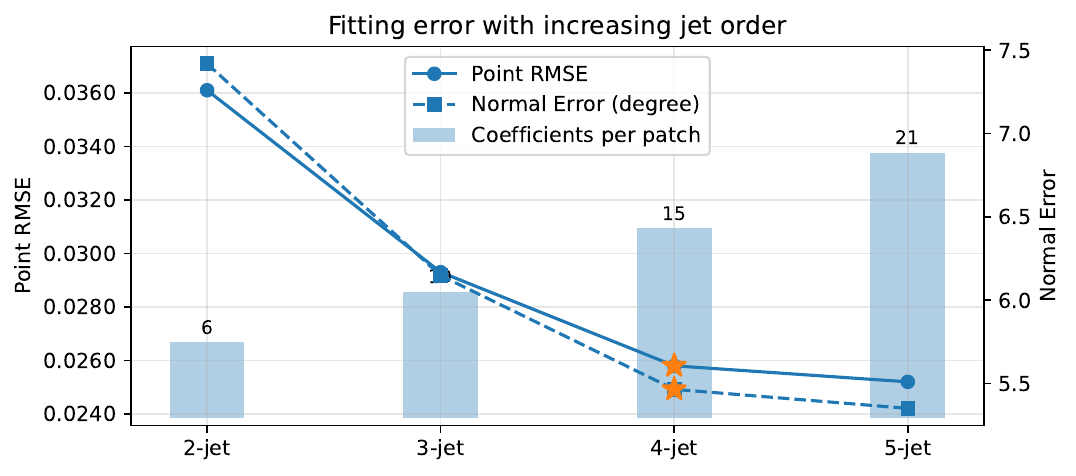}
   \caption{Fitting error with respect to jet order $n$ on CLOTH3D patches.}
   \label{fig:sota_qual}
\end{figure}

\noindent \textbf{Ablation on Jet Order $n$.} To evaluate the fitting performance of PolyFit, we use garment patches from the CLOTH3D validation dataset \cite{Bertiche2020}. We compute its performance from metrics including height RMSE and normal loss, measured in degrees.~\Cref{fig:sota_qual} shows the performance of n-jet fitting on the CLOTH3D dataset. This shows that the 4-jet function is capable of fitting point clouds from garment patches effectively. Therefore, we fix $n=4$, as this setting has been shown to achieve high accuracy on garments with reasonable computational complexity.

~\Cref{tab:polyfit_ablation} indicates that the STN module noticeably enhances the model's fitting accuracy as it re-orients patches to improve their bijectivity, which leads to better jet-fitting.

\begin{table}[htbp]
  \centering
  \resizebox{0.3\textwidth}{!}{
  \begin{tabular}{@{}lccc@{}}
    \toprule
    & Height RMSE &  Normal Diff (°)  \\
    \midrule
    with  & 0.0201 & 5.274\\
    w/o  & 0.0259 & 5.465\\
    \bottomrule
  \end{tabular}
  }
  \vspace{3pt}
  \caption{PolyFit fitting metric, with and w/o STN.}
  \label{tab:polyfit_ablation}
\end{table}

\noindent \textbf{Additional ablation studies.}
We evaluate PolyFit on patches sampled from the CLOTH3D validation split and compare it with PointNet~\cite{Qi2017} and DGCNN~\cite{Wang2019}, two point-based networks that we adapt to regress jet coefficients directly from point clouds.
As summarized in \Cref{tab:why_polyfit}, PolyFit delivers lower geometric error (height RMSE and normal difference) and shorter per-patch inference time than these alternatives.


\begin{table}[htbp]
  \centering
  \resizebox{1\linewidth}{!}{
  \begin{tabular}{@{}lccc@{}}
    \toprule
  Model  & Height RMSE & Normal Diff (°) & Time (ms) \\
    \midrule
    \textbf{PointNet} \cite{Qi2017}  & 0.0309 & 6.936 & 0.0754 \\
    \textbf{DGCNN} \cite{Wang2019}  & 0.0290 & 6.406 & 0.0625\\
    \textbf{PolyFit}  & \textbf{0.0201} &  \textbf{5.274} & \textbf{0.0481}  \\
    \bottomrule
  \end{tabular}
  }
  \vspace{1pt}
  \caption{Comparison with point-based backbones on CLOTH3D validation patches. We report height RMSE, normal-angle error (°), and per-patch inference time (ms).}
  \label{tab:why_polyfit}
\end{table}

We further ablate the family of parametric functions used for training. ~\Cref{tab:abl_study_polyfit} shows that increasing the diversity of parametric functions and augmenting the training set with patches extracted from garment meshes both yield additional accuracy gains.


\begin{table}[htbp]
  \centering
  \resizebox{1\linewidth}{!}{
  \begin{tabular}{@{}lcc@{}}
    \toprule
  Function used for training  & Height RMSE & Normal Diff (°) \\
    \midrule
    Gaussians only  & 0.0248 & 5.485  \\
    4 Families   & 0.0239 & 5.423 \\
    4 Families + garment patches  & 0.0201 & 5.317 \\
    \bottomrule
  \end{tabular}
  }
  \vspace{1pt}
  \caption{Study on different training data for PolyFit. }
  \label{tab:abl_study_polyfit}
\end{table}

\subsection{PolySfT: Implementation details}\label{subsec:polysft_implementation}
\noindent \textbf{Adaptive Window Optimization.} We adopt an adaptive sliding-window optimization strategy with a window size of $W$. Within each window, optimization continues until either the loss fails to improve for a preset number of consecutive iterations (referred to as the \textbf{patience threshold}) relative to the current minimum, or the number of iterations exceeds a certain period (referred to as the \textbf{frame period}). Once either condition is met, we shift the window forward by one frame and initialize the new frame's parameters using those from the previous frame. This method promotes temporal consistency and maximizes optimization efficiency. 

\subsection{PolySfT: Experiments}\label{subsec:polysft_experiments}
\begin{figure}[htbp]
  \centering
\includegraphics[width=1\linewidth]{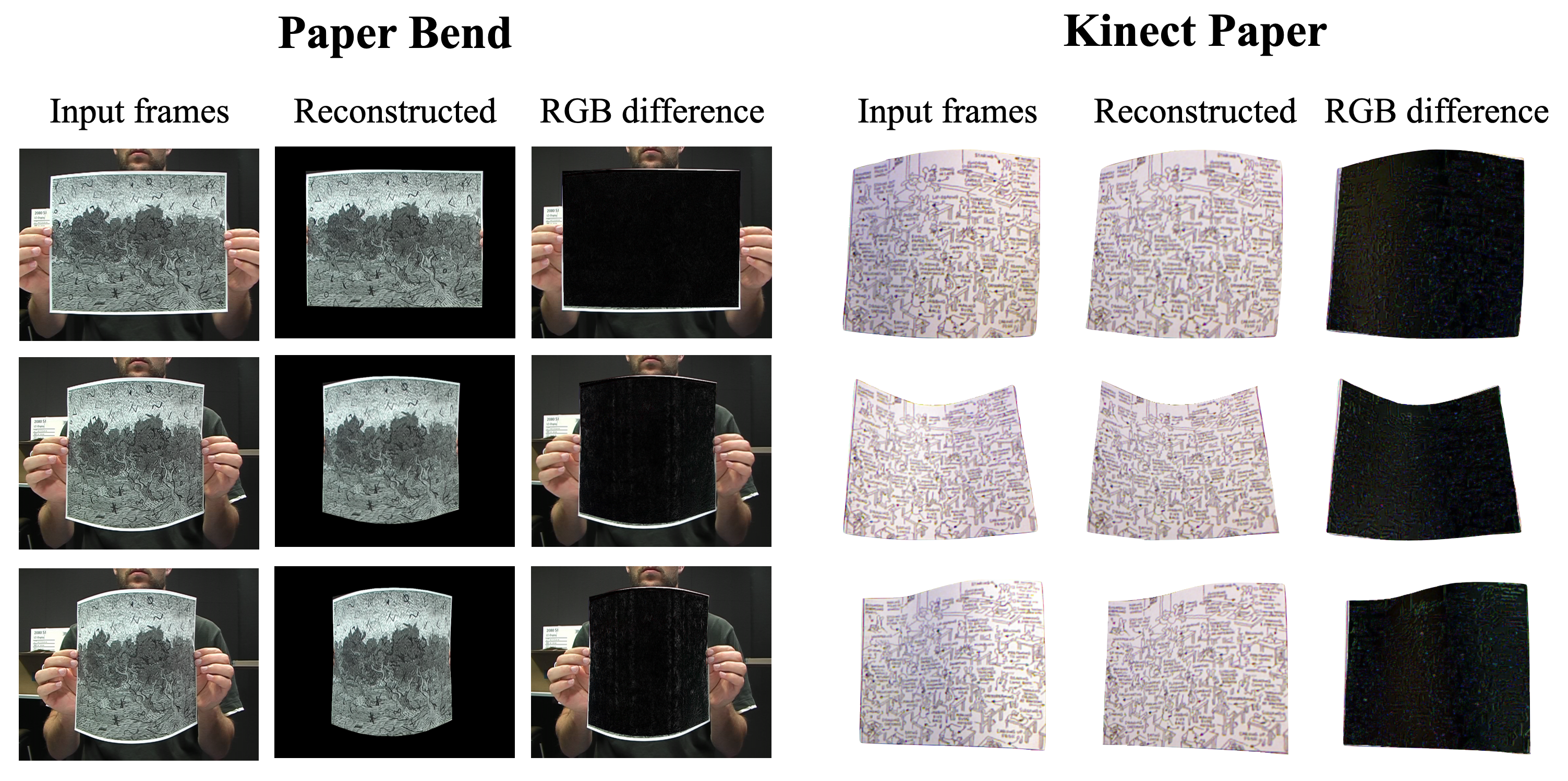}
    \vspace{-5pt}
   \caption{Additional reconstruction results on Paper-Bend and Kinect-Paper.}
   \label{fig:sft_qual}
\end{figure}

\begin{figure}[t]
\centering
\includegraphics[width=1\linewidth, angle=0]{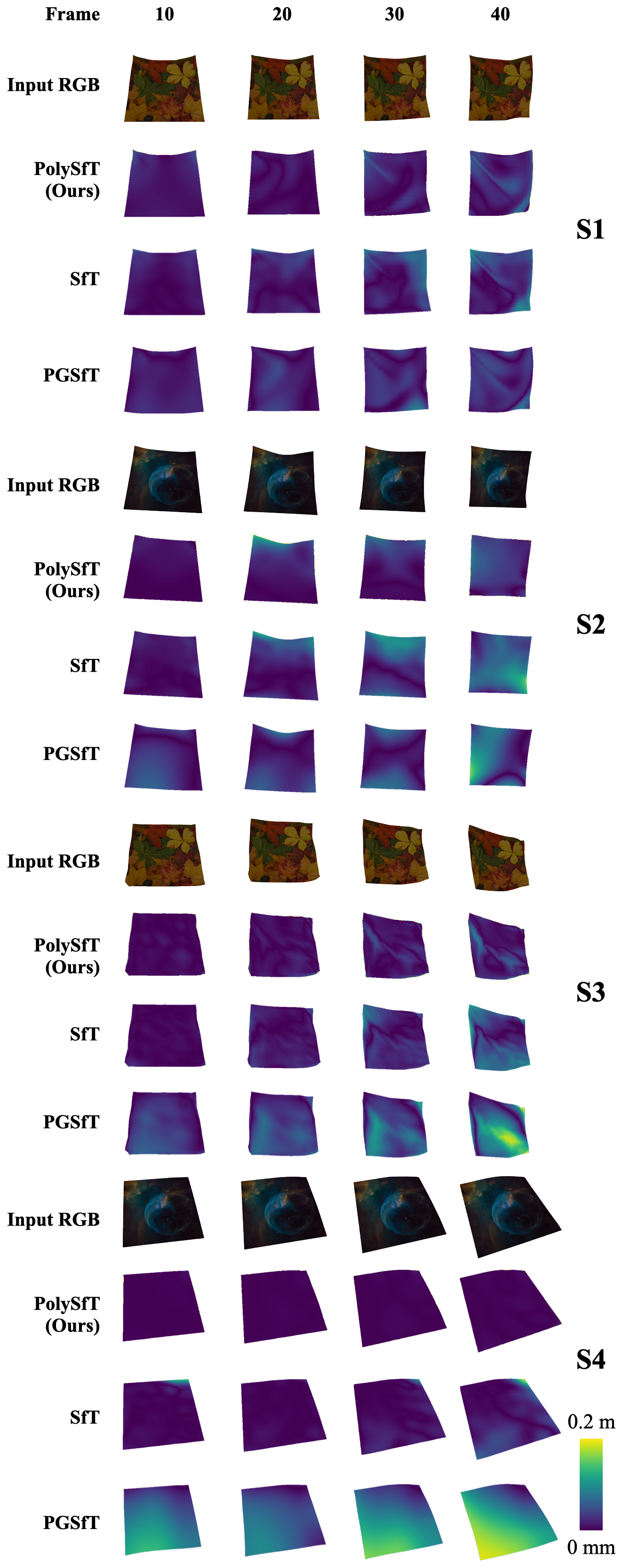}
    \vspace{-5pt}
   \caption{Error map comparison with SOTA methods on $\phi$-SfT Synthetic Dataset, showing frames 10, 20, 30, and 40 from left to right.}
   \label{fig:qual_phi_sft_syn}
\end{figure}

In addition to the quantitative and qualitative results reported in the main paper, we provide further visual results here. \Cref{fig:sft_qual} presents additional qualitative results on the Paper-Bend and Kinect-Paper datasets. Renderings of the reconstructed meshes (second column) closely match the input images (first column), resulting in low per-pixel RGB error maps (third column). \Cref{fig:qual_phi_sft_syn} shows a comparison with SOTA methods on the synthetic dataset provided by \cite{Kairanda2022}.

\begin{figure}[htbp]
  \centering
\includegraphics[width=1\linewidth]{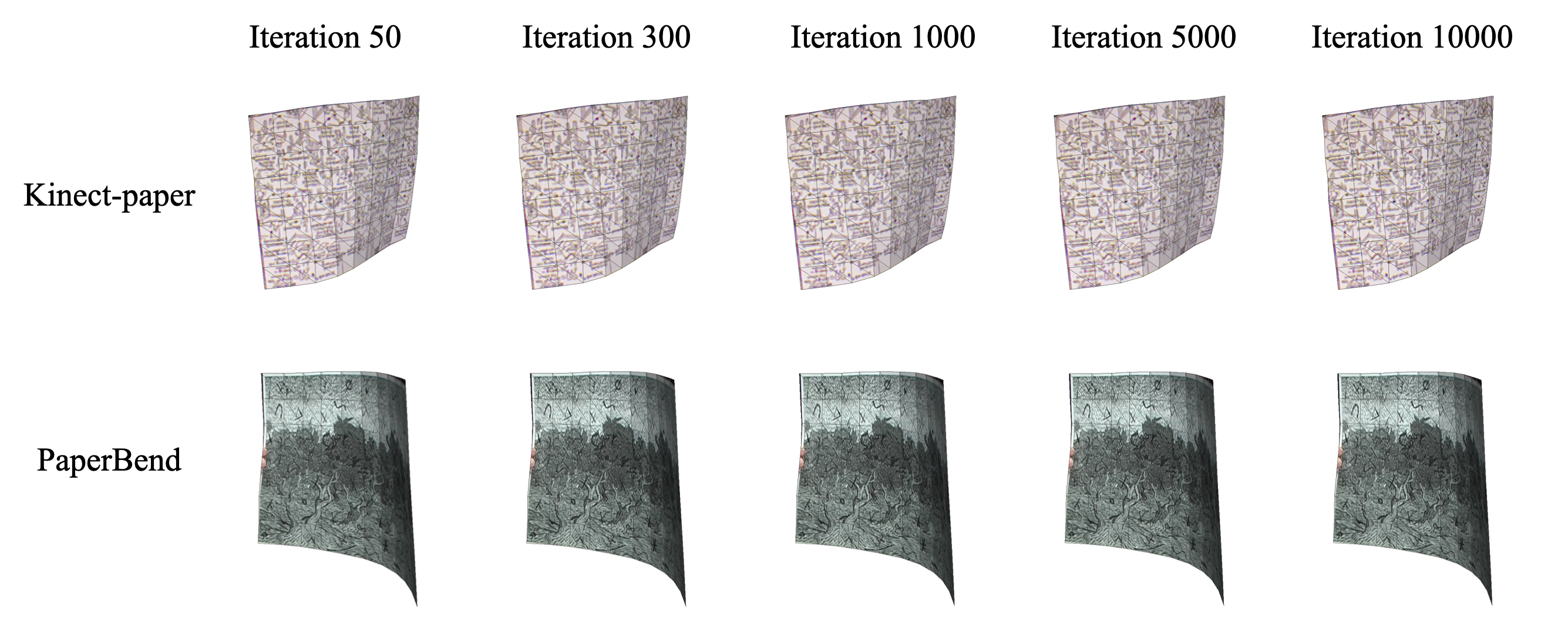}
    \vspace{-5pt}
   \caption{Stability test for Kinect-Paper and Paper-Bend. We show the reconstructed mesh at various iterations.}
   \label{fig:polysft_stability_test}
\end{figure}

\noindent \textbf{Stability test.} We assess PolySfT's stability by running the optimization process for many more iterations than usual. \Cref{fig:polysft_stability_test} displays the reconstructed meshes for two scenes with different motion patterns at 50 iterations, 300 iterations (the average evaluation point), and extended runs at 1000 and 5000 iterations. The results demonstrate that the mesh reliably tracks the intended motion, with only minimal changes beyond the typical iteration threshold. Moreover, initializing from previous frames provides a robust starting point for the current frame. Experiments are conducted on a single NVIDIA V100 GPU.

\subsection{OneFit: Network and training details}\label{sec:add_details}
In this section, we provide details of the OneFit architecture and training setup. In the Dynamic encoder, different from~\cite{Bertiche2022}, the Gated Recurrent Unit (GRU) layers are initialized with random hidden states. The body feature extractor is implemented using a five-layer multilayer perceptron (MLP) with LeakyReLU activation between the layers. Each layer contains 256 nodes, with the exception of the final layer. 

The decoder consists of four fully connected layers, each with dimensions of 512, 512, 512, and 256, respectively. This is followed by three prediction heads for jet coefficients, translation and scale, each implemented as three fully connected layers with dimensions 128 and 64, ending with a final output layer.

Finally, to maximize parallel computation on GPUs, the batch size for each garment is dynamically determined based on the number of patches using the following equation: $bs=\frac{20,000}{\text{number of patches}}$.

\subsection{OneFit: Garment preprocessing}\label{subsec:Garment-preprocessing}
We describe the preprocessing used to align CLOTH3D garments to the average SMPL body in T-pose.

\noindent \textbf{T-pose average shape conversion.} In CLOTH3D, garments are posed with legs slightly apart, differing from the standard SMPL T-pose on which skinning weights are defined. In addition, the dataset is fitted on different body shapes. To evaluate garments from CLOTH3D with OneFit, we first preprocess each garment to align it with the average SMPL body in standard T-pose. Specifically, for each garment vertex we query the closest body vertex and displace it according to the difference between the original and standard body shapes. A single iteration of Laplacian surface smoothing is then applied to remove local artifacts. For loose garments such as dresses and skirts, which do not adhere closely to the legs, we only correct the position in terms of shape difference without enforcing pose alignment.

\subsection{OneFit: Experiments}\label{subsec:onefit_exp}

\begin{figure}[htbp]
  \centering
\includegraphics[width=1\linewidth]{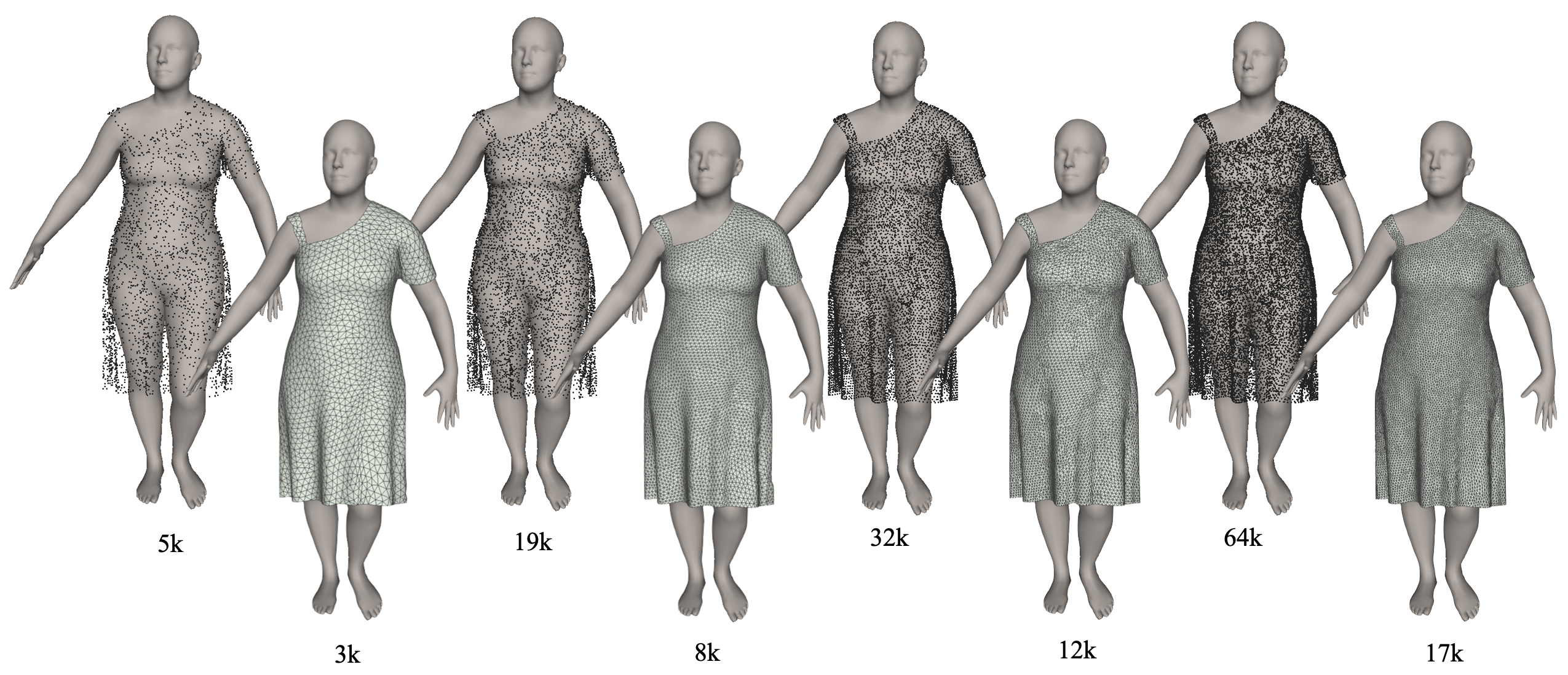}
\vspace{-3pt}
   \caption{\textbf{OneFit} drapings with different mesh resolutions obtained within a similar inference time.}
   \label{fig:resolution}
\end{figure}

\end{document}